\documentclass[times, preprint, 10pt]{elsarticle}

\usepackage{amssymb}
\usepackage{amsmath}
\usepackage{graphicx}
\usepackage{multirow}
\usepackage{array}
\usepackage{booktabs} 
\usepackage{xcolor}
\usepackage{hyperref}
\usepackage{setspace}

\journal{Pattern Recognition}

\begin{document}

\begin{frontmatter}

\title{On the Design Fundamentals of Diffusion Models: \\A Survey}

\author[add1]{Ziyi~Chang}
\ead{ziyi.chang@durham.ac.uk}
\author[add1]{George~A.~Koulieris}
\ead{george.koulieris@durham.ac.uk}
\author[add2]{Hyung~Jin~Chang}
\ead{h.j.chang@bham.ac.uk}
\author[add1]{Hubert~P.~H.~Shum\corref{cor1}}
\cortext[cor1]{Corresponding author}
\ead{hubert.shum@durham.ac.uk}

\address[add1]{Department of Computer Science, Durham University, Durham, DH1 3LE, UK}
\address[add2]{School of Computer Science, University of Birmingham, Birmingham, B15 2TT, UK}

\begin{abstract}
Diffusion models are learning pattern-learning systems to model and sample from data distributions with three functional components namely the forward process, the reverse process, and the sampling process. The components of diffusion models have gained significant attention with many design factors being considered in common practice. Existing reviews have primarily focused on higher-level solutions, covering less on the design fundamentals of components. This study seeks to address this gap by providing a comprehensive and coherent review of seminal designable factors within each functional component of diffusion models. This provides a finer-grained perspective of diffusion models, benefiting future studies in the analysis of individual components, the design factors for different purposes, and the implementation of diffusion models.
\end{abstract}


\end{frontmatter}



\section{Introduction}
Diffusion models, as a learning system, consist of three functional components, i.e., the forward, reverse, and sampling processes. The generic pipeline of diffusion models \cite{sohl2015deep} involves forward and reverse processes to learn a data distribution, and a sampling process to generate novel data that follow such a distribution. Three components work together to achieve the functionality of diffusion models \cite{karras2022elucidating}, i.e., the ability to model and sample from data distributions. The forward process is expected to perturb training data by adding noise while the reverse process is expected to remove the aforementioned perturbation via learning a neural network. After optimization, the sampling process is expected to generate novel data that follow the distribution.

Designing the three functional components involves consideration of different major factors and design purposes. To perturb training data, the schedule of noise and the type of noise are two major factors that need to be considered when a forward process is designed. Meanwhile, the terminal point and diffusion space are also required to specify where to stop and where to manipulate data in a forward process. The forward process needs to perturb the training data by adding noise at each timestep. For a reverse process to remove the added noise, architectures, parameterizations, and optimizations of a neural network become major factors to be considered for the reverse process. Architectures specify how to learn denoising while parameterizations concern what to learn and predict. Optimizations allow different emphases on the information that a neural network should focus on more. The sampling process works after optimization, and the controllability and speed become two major factors to be considered. Controllability constrains the generation to obtain data of users' interest while the speed factor accelerates generation without significant quality degradation. While there could be a large number of factors in each functional component, our survey only focuses on major ones that are transferable and usually considered in common practice.

\begin{figure}[htbp]
  \label{fig:overview}
  \centering
  \includegraphics[width=0.9\linewidth]{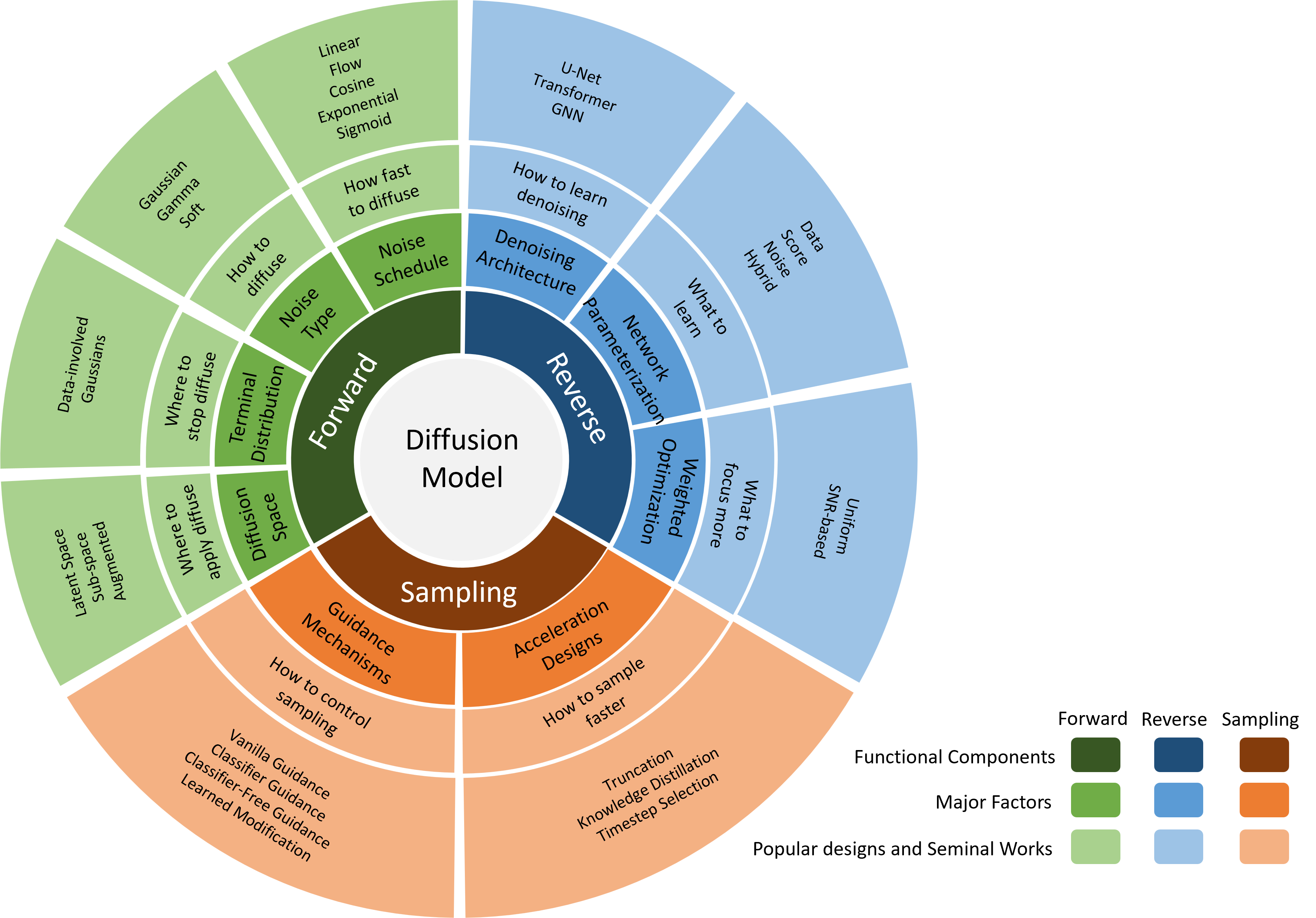}
  \caption{The hierarchical overview of diffusion models. The forward process, the reverse process, and the sampling process are three functional components. Major factors comprise each component. Popular designs and seminal works are presented.}
\end{figure}

Most existing survey papers on diffusion models focus on a particular application area while ours hierarchically organizes the design fundamentals in a diffusion model. With the recent prosperity in applications of diffusion models, previous survey papers mostly focus on collecting application cases in various domains and data structures. These domains include natural language processing \cite{zhu2023diffusion}, computer vision-related tasks \cite{wang2025adv}, medical analysis \cite{kazerouni2022diffusion}, natural science \cite{zhang2023surveygraph}, time series \cite{lin2023diffusion}, recommendation \cite{jiangzhou2024dgrm}, personalization \cite{zhang2024survey}, memorization \cite{wang2024replication}, and etc. They also cover different types of data, including image \cite{kim2024depth,wang2025lldiffusion}, text \cite{li2023diffusion}, video \cite{xing2023survey}, audio \cite{zhang2023audio}, etc. However, they are domain-specific and application-driven, and thus lead to restricted insights into this whole area. Some recent surveys \cite{cao2022survey,yang2022diffusionsurvey,chen2024overview} are organized by specific problems, which may hinder a comprehensive understanding. In contrast, our survey adopts a design-centric taxonomy, providing building blocks to facilitate straightforward implementation.

Our paper treats diffusion models as a learning system, discusses the system hierarchically, and mainly focuses on seminal designable factors within each component, as shown in Fig. \ref{fig:overview}. This breakdown is aligned with the functionality of diffusion models and the intuition of getting to know a system. Therefore, our survey benefits both beginners who want to get fundamental knowledge from seminal works in this area and professionals who want to hierarchically understand critical factors when designing their advanced diffusion models.

The following are questions to answer via our literature review of diffusion models:
\begin{enumerate}
    \item What are the \textbf{functional components} in a diffusion model?
    \item What are the \textbf{major factors} that comprise each component?
    \item What are the \textbf{popular designs} and \textbf{seminal works} in each factor?
\end{enumerate}

We organize our survey by answering the above questions to build a hierarchical view of diffusion models. Section \ref{sec:preliminaries} introduces the generic pipeline of diffusion models, including the three functional components and two popular formulations. Sections \ref{sec:forward}, \ref{sec:reverse}, and \ref{sec:sampling} respectively review the major factors, and popular designs and seminal works of each functional component, i.e., the forward process, the reverse process, and the sampling process. Section \ref{sec:future} provides overall insights on the future trends with respect to this field, and Section \ref{sec:conclusion} gives a brief conclusion on diffusion models.

\section{The Generic Pipeline}\label{sec:preliminaries}

This survey uses commonly-used notations and terminologies in existing papers and represents concepts with figures whose legends are defined in Table \ref{tab:legends}.

\begin{table}[htbp]
\small
\centering
\begin{tabular}{cm{0.75\columnwidth}<{\centering}}
\hline
Notation & Legends \\
\hline
\begin{minipage}{8mm}\includegraphics[width=8mm, height=4mm]{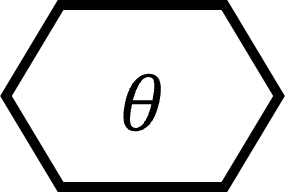}\end{minipage}
& Trainable network with parameters $\theta$\\
\begin{minipage}{8mm}\includegraphics[width=8mm, height=4mm]{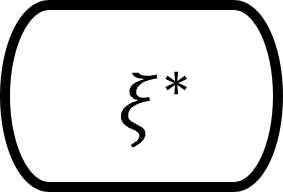}\end{minipage}
& Fixed network with parameters $\xi^*$\\
\begin{minipage}{5mm}\includegraphics[width=5mm, height=5mm]{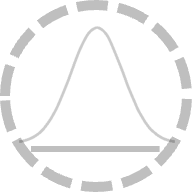}\end{minipage}\quad\begin{minipage}{10mm}\includegraphics[width=10mm, height=5mm]{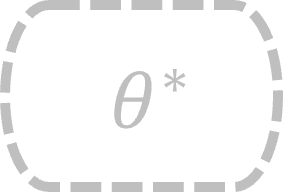}\end{minipage}
& Component not in use\\
\begin{minipage}{10mm}\includegraphics[width=10mm, height=4mm]{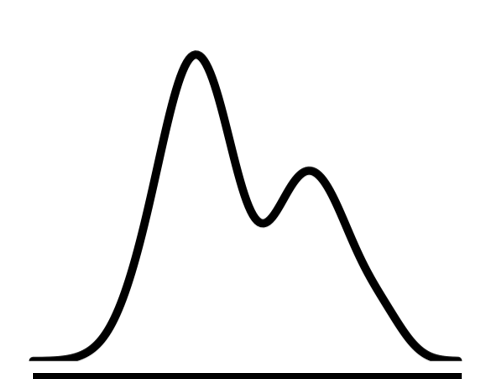}\end{minipage}\quad\begin{minipage}{10mm}\includegraphics[width=10mm, height=8mm]{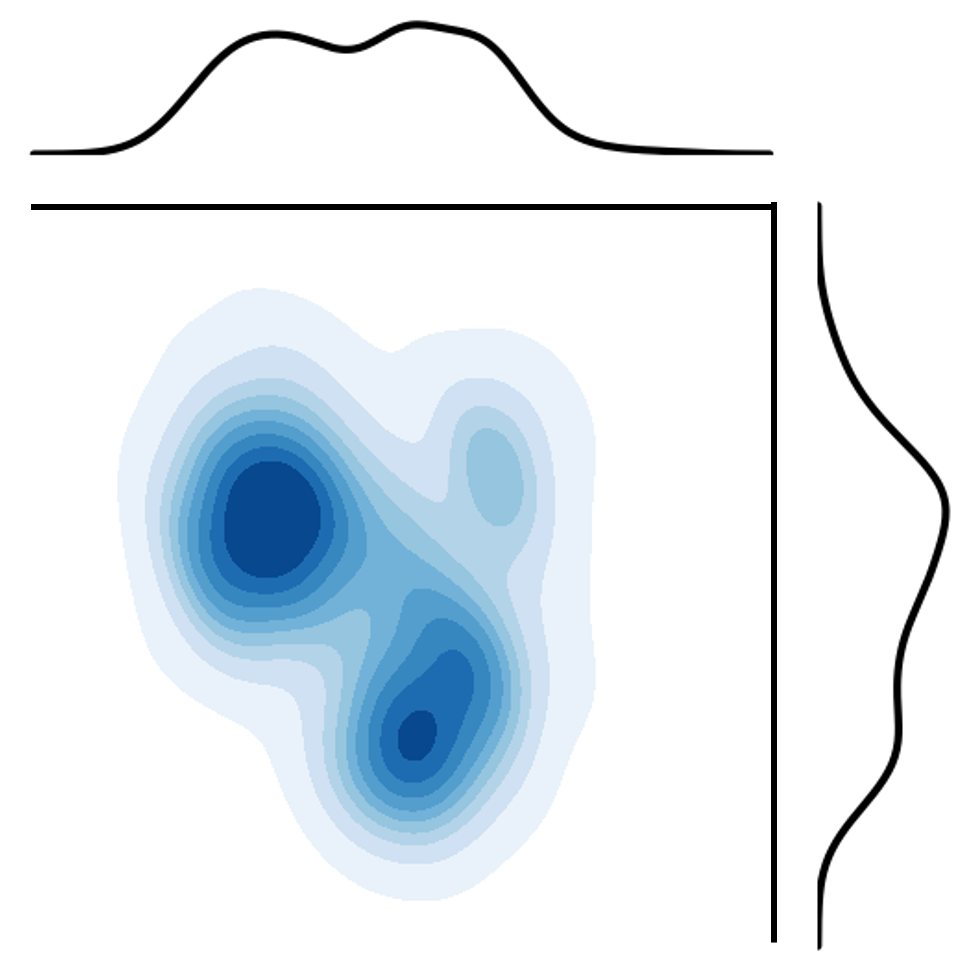}\end{minipage}
& Data distributions\\
\begin{minipage}{5mm}\includegraphics[width=5mm, height=5mm]{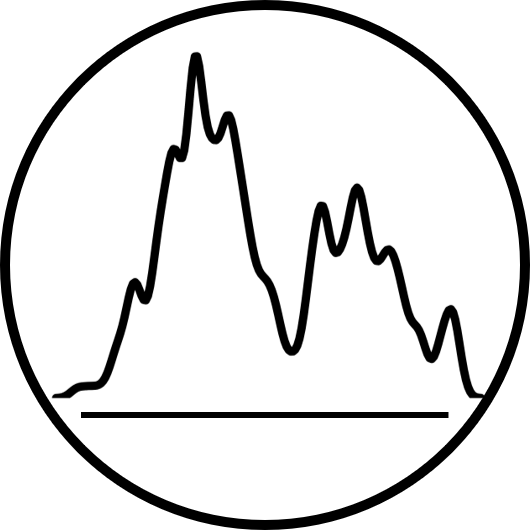}\end{minipage}
& The distribution at a timestep\\
\begin{minipage}{5mm}\includegraphics[width=5mm, height=5mm]{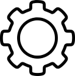}\end{minipage}
& Condition $c$\\
\begin{minipage}{5.5mm}\includegraphics[width=5.5mm, height=5.5mm]{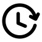}\end{minipage}
& Timestep $t$\\
\begin{minipage}{5mm}\includegraphics[width=5mm, height=5mm]{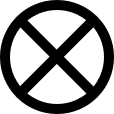}\end{minipage}
& Combination, e.g., Addition.\\
\begin{minipage}{5mm}\includegraphics[width=5mm, height=5mm]{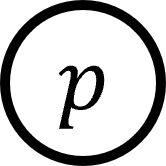}\end{minipage}
& With probability $p$ for dropping out\\
\hline
\end{tabular}
\caption{Figure legends.}
\label{tab:legends}
\end{table}

\subsection{Three Functional Components}\label{sec:pipeline}

\begin{figure}[tbph]
  \centering
  \includegraphics[width=0.95\linewidth]{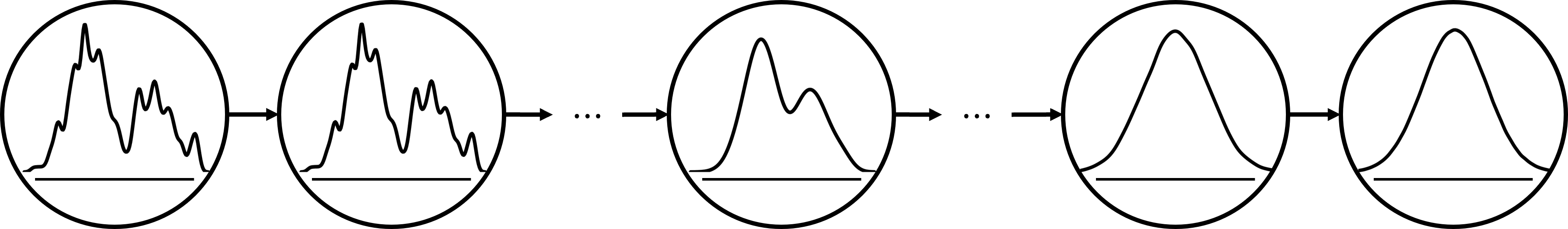}
  \caption{\label{fig:forward_process}%
          The forward process perturbs the original unknown distribution by gradually adding noise to a given set of data samples through a chain of distribution transitions with multiple time steps. Each time step of the chain is denoted by a circle.
          }
\end{figure}

\textbf{The forward process} perturbs a training sample $x_0$ to $\{x_t\}_{t=1}^T$ as the timestep $t$ increases, as shown in Fig. \ref{fig:forward_process}. A forward transition $p(x_t|x_{t-1})$ describes such a perturbation where a small amount of noise $\epsilon_t$ is added between two timesteps. In other words, as the forward process moves on the chain, more and more noise is added through $p(x_t|x_{t-1})$ and the perturbed sample $x_t$ becomes noisier and noisier. Through multiple timesteps, the original distribution $p(x_0)$ is eventually perturbed to a tractable terminal distribution $p(x_T)$, which is usually full of noise. Since only noise is added through the chain, the forward process does not have any trainable parameters. In particular, the forward process is represented as a chain of forward transitions:
\begin{equation}\label{eq:forward_process}
        p\left(x_{1:T}|x_0\right):=\prod_{t=1}^Tp\left(x_t|x_{t-1}\right),
\end{equation}
where $t$ is the timestep, $T$ is the total number of timesteps, $x_0$ is a training sample at $t=0$ and is then perturbed to be $x_T$ after $T$ timesteps, and $p\left(x_t|x_{t-1}\right)$ is a forward distribution transition between two consecutive time steps.

\begin{figure}[tbph]
  \centering
  \includegraphics[width=0.95\linewidth]{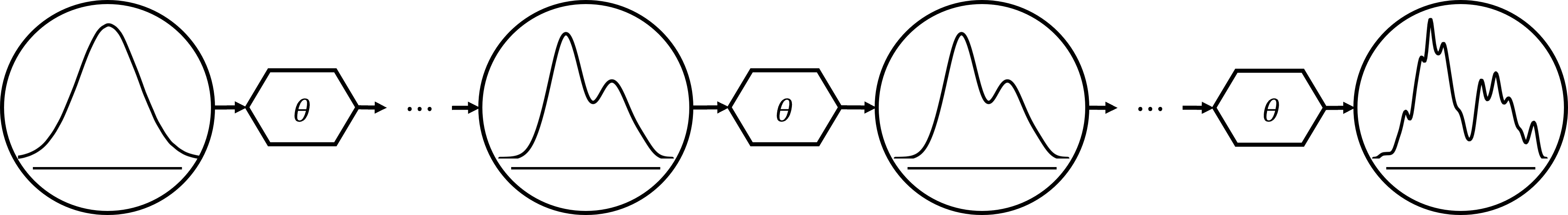}
  \caption{\label{fig:reverse_process}%
          The reverse process trains a neural network $\theta$ to recursively remove the noise that has been previously added by the forward process.
          }
\end{figure}

\textbf{The reverse process} trains a denoising network to recursively remove the noise, as shown in Fig. \ref{fig:reverse_process}. A denoising network is trained to iteratively remove the noise between two consecutive timesteps. The reverse process moves backwards on the multi-step chain as $t$ decreases from $T$ to $0$. Such iterative noise removal is termed as the reverse transition $p_\theta(x_{t-1}|x_t)$, which is approximated by optimizing the trainable parameters $\theta$ in the denoising network. 
In particular, the reverse process is formulated as a chain of reverse transitions:
\begin{equation}\label{eq:reverse_process}
    p_\theta\left(x_{0:T}\right):=p\left(x_T\right)\prod_{t=1}^Tp_\theta\left(x_{t-1}|x_t\right),
\end{equation}
where $\theta$ is the parameters of the denoising network and $p_\theta(x_{t-1}|x_t)$ is the reverse distribution transition. In particular, the reverse process is usually parameterized as:
\begin{equation}\label{eq:reverse_transition}
    p_\theta\left(x_{t-1}|x_t\right):=\mathcal{N}\left(x_{t-1};\mu_\theta\left(x_t,t\right),\Sigma_\theta\left(x_t,t\right)\right),
\end{equation}
where $\mu_\theta(x_t,t)$ and $\Sigma_\theta(x_t,t)$ are, respectively, the Gaussian mean and variance to be estimated by the network $\theta$.

The denoising network is trained by the standard variational bound:
\begin{equation}
    \begin{aligned}\label{eq:likelihood_objective}
        L=\mathbb{E}\Bigg[&\underbrace{D_{KL}\Bigl(p\left(x_T|x_0\right)||p\left(x_T\right)\Bigr)}_{\text{prior matching term}}\\
        &+\underbrace{\sum_{t\geq1}D_{KL}\Bigl(p\left(x_{t-1}|x_t,x_0\right)||p_\theta\left(x_{t-1}|x_t\right)\Bigr)}_{\text{denoising matching term}}\\
        &-\underbrace{\log p_\theta\left(x_0|x_1\right)}_{\text{reconstruction term}}\Bigg],\\
    \end{aligned}
\end{equation}
where $D_{KL}(\cdot\|\cdot)$ is the Kullback–Leibler (KL) divergence to compute the difference between two distributions. The prior matching term is minimized as the final distribution becomes Gaussian after a sufficiently large $T$. The reconstruction term can be approximated using a Monte Carlo estimate, and training primarily focuses on the denoising matching term. Overall, minimization of  the objective $L$ is to reduce the discrepancy between $p_\theta(x_0)$ and $p(x_0)$.

\begin{figure}[tbph]
  \centering
  \includegraphics[width=0.95\linewidth]{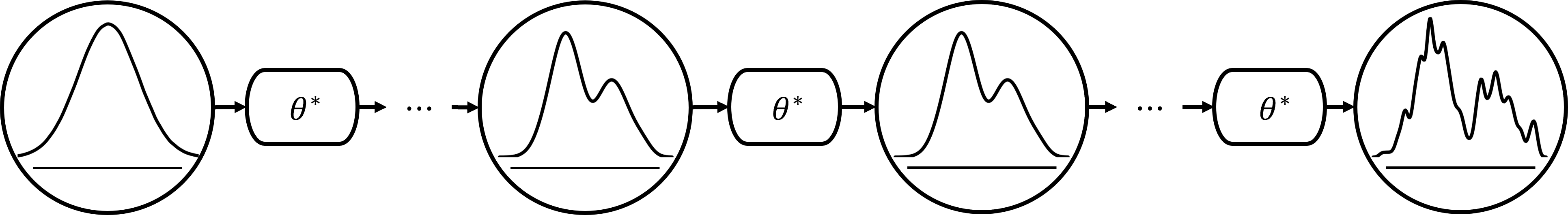}
  \caption{\label{fig:sampling_procedure}%
          The sampling process uses the trained denoising network $\theta^*$ and usually follows the same transitions as the reverse process.
          }
\end{figure}

\textbf{The sampling process} leverages the optimized denoising network $\theta^*$ to generate novel data $x_0^*$, as illustrated in Fig. \ref{fig:sampling_procedure}. It moves backwards on the chain to recursively apply the optimized network $\theta^*$. Concretely, it firstly obtains a sample $x_T$ from the terminal distribution $p(x_T)$ and then uses the trained network to iteratively remove noise by the sampling transition $p_{\theta^*}(x_{t-1}|x_t)$. Through a chain of such transitions, it finally generates new data $\hat{x}_0\sim p_{\theta^*}(x_0)\approx p(x_0)$. In particular, the sampling process is defined as a chain of sampling transitions:
\begin{equation}
    p_{\theta^*}\left(x_{0:T}\right):=p\left(x_T\right)\prod_{t=1}^Tp_{\theta^*}\left(x_{t-1}|x_t\right),
\end{equation}
where $\theta^*$ represents the optimized parameters of the denoising network, $p(x_T)$ is the terminal distribution, and $p_{\theta^*}(x_{t-1}|x_t)$ is the sampling transition.

\subsection{Discrete and Continuous Formulations}

To reflect the development of diffusion models, we organize diffusion models by two popular formulations, i.e., discrete and continuous timesteps. To keep our survey simple to understand, especially for beginners, we present and discuss most of the fundamental designs under the discrete-time framework. These choices on the discrete-time framework are generally applicable to the continuous-time framework.

\subsubsection{The Discrete Formulation}\label{sec:discrete_formulation}

Initially motivated by unsupervised learning, diffusion models are formulated with discrete timesteps. Regarding the discrete formulation, the denoising diffusion probabilistic model (DDPM) \cite{ho2020denoising} is a popular configuration of such formulated diffusion models. It is straightforward to define, efficient to train, and capable of achieving high quality and high diversity in the generated samples \cite{dhariwal2021diffusion}. 

Concretely, the forward transition in DDPM is defined to add isotropic Gaussian noise $\epsilon_t\sim\mathcal{N}(0,I)$:
\begin{equation}\label{eq:ddpm_forward_process}
    p(x_t|x_{t-1}):=\mathcal{N}(x_t;\sqrt{1-\beta_t}x_{t-1},\beta_tI),
\end{equation}
where $\beta_t$ is the noise schedule, which is a hyper-parameter to control the amount of noise to be added in each timestep. As all forward transitions are Gaussian, the forward process in DDPM is simplified as:
\begin{equation}\label{eq:ddpm_simplified_forward}
    p(x_t|x_0):=\mathcal{N}(x_t;\sqrt{\bar{\alpha}_t}x_0,(1-\bar{\alpha}_t)I),
\end{equation}
where $\bar{\alpha}_t$ is defined as $\bar{\alpha}_t=\prod_{s=1}^t\alpha_s$ and $\alpha_t=1-\beta_t$. In theory, $\bar{\alpha}_t$ has a similar effect with $\beta_t$ in Eq. (\ref{eq:ddpm_forward_process}).

The reverse process has the same functional form as the forward process \cite{sohl2015deep}. In DDPM configuration, the transition Eq. (\ref{eq:reverse_transition}) in the reverse process is formulated as:
\begin{equation}\label{eq:ddpm_reverse_process}
    p_\theta(x_{t-1}|x_t):=\mathcal{N}(x_{t-1};\frac{1}{\sqrt{\alpha_t}}(x_t-\frac{1-\alpha_t}{\sqrt{1-\bar{\alpha}_t}}\epsilon_\theta(x_t,t)),\beta_tI),
\end{equation}
where the variance $\Sigma_\theta(x_t,t)$ in Eq. (\ref{eq:reverse_transition}) is empirically fixed as the noise schedule $\beta_t$, and $\mu_\theta(x_t,t)$ is reparametrized by the noise prediction $\epsilon_\theta\left(x_t,t\right)$. Accordingly, the training objective defined in Eq. (\ref{eq:likelihood_objective}) is also simplified as:
\begin{equation}
    L=\mathbb{E}_{x_t,t}\left[\left\|\epsilon_t-\epsilon_\theta\left(x_t,t\right)\right\|_2^2\right].
\end{equation}
The intuition behind the derivation is two-fold. First, all distributions involved in Eq.~\ref{eq:likelihood_objective} are Gaussians. Second, with Eq.~\ref{eq:ddpm_forward_process} and Eq.~\ref{eq:ddpm_reverse_process}, the KL divergence is simplified to be only dependent on the mean that is parameterized by the predicted noise $\epsilon_t$. Finally, the sampling process obtains $x_T\sim p(x_T)$, and applies $p_{\theta^*}(x_{t-1}|x_t)$ to generate $\hat{x}_0$.

\subsubsection{The Continuous Formulation}\label{sec:continuous_formulation}
Focusing on the dynamics of diffusion models, continuous formulation is proposed to analyze the complex dynamics and also integrate the domain knowledge of score matching. The continuous formulation manipulates data distributions in continuous time. Noise is added in an infinitesimal interval between timesteps. Therefore, a differential equation (DE) is adopted in such formulated diffusion models to describe changes in continuous timesteps. Furthermore, \cite{song2020score} unifies all diffusion models with differential equations. Flow matching or stochastic interpolants \cite{lipman2023flow,albergo2023building,albergo2023stochastic,liu2023flow} is one of the popular approaches to improve the dynamics of diffusion models, which is often presented using continuous formulation.

Concretely, the forward transition to add noise is formulated as a forward SDE:
\begin{equation}
    dx=f(x,t)dt+g(t)dw,
\end{equation}
where $w$ is the standard Wiener process and accounts for noise in the forward transition, and $f(x,t)$ and $g(t)$ are the drift and diffusion coefficients to account for the mean and variance in the forward transitions, respectively.

At the same time. a reverse SDE for the reverse transition is also determined by these coefficients. Specifically, the reverse SDE is:
\begin{equation}\label{eq:sde_reverse_process}
    dx=\left[f\left(x,t\right)-g^2\left(t\right)s_\theta\left(x_t,t\right)\right]dt+g\left(t\right)dw,
\end{equation}
where the output of the denoising network $s_\theta(x_t,t)=\nabla_x\log p(x_t)$ is the score.
Likewise, $\left[f(x,t)-g^2(t)s_\theta(x_t,t)\right]$ and $g(t)$ account for the mean and the variance in Eq. (\ref{eq:reverse_transition}). The training objective is defined as:
\begin{equation}\label{eq:learning_objective}
    L=\mathbb{E}_t\left[\lambda\left(t\right)\mathbb{E}_{x_0}\mathbb{E}_{x_t|x_0}\left[\left\|s_\theta\left(x_t,t\right)-\nabla_x\log p(x_t|x_0)\right\|_2^2\right]\right],
\end{equation}
where $\lambda(t)$ is the weighting function. Finally, the sampling process obtains $x_T$, and applies the trained network $\theta^*$ to generate novel data.

The ODE-based formulation, derived from a deterministic process with the same marginal densities $\{p(x_t)\}_{t=0}^T$ as the SDE, is detailed in \cite{song2020score} and expressed as:
\begin{equation}
    dx=\left[f\left(x,t\right)-\frac{1}{2}g^2\left(t\right)s_\theta\left(x_t,t\right)\right]dt,
\end{equation}
Its optimization objective matches Eq.~\ref{eq:learning_objective}. Compared to SDE-based formulations, ODE-based approaches enable exact likelihood computation and provide deterministic latent representations, which are advantageous for editing tasks and efficient sampling.

\subsection{Evaluation}
The evaluation of diffusion models encompasses three dimensions: distribution, individual, and human aspects, the first two of which are often combined with domain knowledge. From a distribution perspective, metrics such as the Frechet Inception Distance \cite{heusel2017gans}, assess the distributional similarity to evaluate generation quality. The individual aspect usually involves paired evaluations such as the latent alignment \cite{saharia2022photorealistic} or self-contained evaluations such as physics constraints \cite{tevet2022human} of generated data. Domain-specific knowledge is essential to design evaluation metrics for different fields such as medical analysis and meteorology. However, these quantitative measures may not align with human perceptual judgments \cite{stein2024exposing}. Human-centric evaluations, such as user studies, are necessarily employed to provide subjective assessments of performance. Recently, human preferences \cite{wu2023human, wu2023human2,xu2023imagereward} have been leveraged for better alignment. With these three evaluation dimensions, diffusion models are rigorously assessed and comparable with other models.

\section{The Forward Process}\label{sec:forward}
The forward process is crucial for the success of diffusion models, as it reduces the complexity of the generation task by positive-incentive noise \cite{li2022positive}. It perturbs data by scheduled noise, and results in stable noisy augmentations of data \cite{zhang2024data} that bridge data and pure noise distributions. These augmentations later facilitate diffusion models to learn the gradual multi-step generation, thereby reducing the task entropy. Moreover, \cite{zhang2023variational} demonstrates that the integration of positive-incentive noise consistently enhances performance from a variational perspective.

\subsection{The Noise Schedule}\label{sec:noise_schedule}

A suitable schedule balances exploration and exploitation \cite{hang2024improved}. Exploration, defined as a model’s ability to generalize to unseen data, requires an adequate level of noise while excessive noise can lead to suboptimal convergence. Conversely, exploitation, where the model effectively fits the training data, is achieved with minimal noise while insufficient noise undermines generalization.

\begin{table}[tbph]
\centering
\begin{tabular}{m{0.25\columnwidth}<{\centering}m{0.7\columnwidth}<{\centering}}
\hline
Noise Schedule         & Visualization \\
\hline
Linear \cite{ho2020denoising}&\begin{minipage}{64mm}
      \includegraphics[width=64mm, height=8mm]{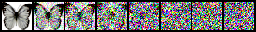}
    \end{minipage} \\
Flow \cite{chen2023importance}&\begin{minipage}{64mm}
      \includegraphics[width=64mm, height=8mm]{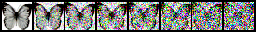}
    \end{minipage} \\
Cosine \cite{nichol2021improved}&\begin{minipage}{64mm}
      \includegraphics[width=64mm, height=8mm]{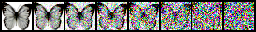}
    \end{minipage} \\
Exponential \cite{song2019generative}&\begin{minipage}{64mm}
      \includegraphics[width=64mm, height=8mm]{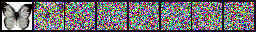}
    \end{minipage} \\
Sigmoid \cite{jabri2022scalable}&\begin{minipage}{64mm}
      \includegraphics[width=64mm, height=8mm]{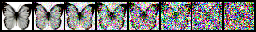}
    \end{minipage} \\
\hline
\end{tabular}
\caption{Illustrations of typical manually-designed noise schedules.}
\label{tab:noise_schedule}
\end{table}

The noise schedule can be either learned by a network or empirically designed using mathematical formulations. Schedules are treated as a parameter to be learned jointly with other parameters \cite{sahoo2024diffusion}. Manually designed noise schedules are formulated with a wide variety of mathematical heuristics. A linear schedule \cite{ho2020denoising} has been initially proposed. For faster perturbation, an exponential schedule \cite{song2019generative} is proposed for better exploration. In contrast, schedules such as cosine \cite{nichol2021improved}, sigmoid \cite{jabri2022scalable}, and rectified flow \cite{liu2023flow} and its variants \cite{ma2024sit,esser2024scaling} are proposed for smoother perturbation speed. Table \ref{tab:noise_schedule} shows several examples of designed noise schedules.

Learning schedules enable models to adaptively optimize noise distributions but may lead to overfitting and reduced generalization \cite{sahoo2024diffusion}. In contrast, predefined schedules avoid adding parameters and offer greater interpretability. They also support different perturbation speeds for trading-off exploration and exploitation. However, they often require manual tuning for a new task or a new dataset \cite{chen2023importance}.

\subsection{The Noise Type}\label{sec:noise_type}

The selection of the noise type plays a crucial role in diffusion models, influencing distribution approximation and the overall expressiveness of the model. An appropriate noise type enhances the capacity to accurately fit the perturbed distributions at various timesteps \cite{nachmani2021denoising}. Additionally, different noise types offer varying degrees of freedom \cite{nachmani2021non}, providing flexibility in modeling complex data distributions.

\begin{table}[tbph]
\centering
\begin{tabular}{m{0.25\columnwidth}<{\centering}m{0.7\columnwidth}<{\centering}}
\hline
Noise Type                   &  Visualization \\
\hline
Gaussian \cite{song2019generative,ho2020denoising}      &\begin{minipage}{64mm}
      \includegraphics[width=64mm, height=8mm]{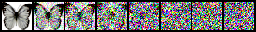}
    \end{minipage}      \\
Gamma \cite{nachmani2021denoising,nachmani2021non}      &\begin{minipage}{64mm}
      \includegraphics[width=64mm, height=8mm]{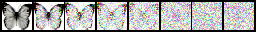}
    \end{minipage}      \\
Soft \cite{bansal2022cold,daras2022soft}      &\begin{minipage}{64mm}
      \includegraphics[width=64mm, height=8mm]{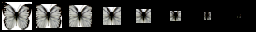}
    \end{minipage}      \\
\hline
\end{tabular}%
\caption{Comparison of several streams of noise types.}
\label{tab:noise_type}
\end{table}

Different noise types have been developed based on empirical experiments. Isotropic Gaussian noise \cite{ho2020denoising} is commonly used for its simplicity and compatibility. It allows for analytical solutions by taking advantage of the additivity of Gaussian distributions and simplifying the calculation of KL divergence in Eq.~\ref{eq:likelihood_objective}. Several variants of isotropic Gaussian noise, such as mixture of Gaussian noise \cite{nachmani2021non} and non-isotropic Gaussian noise \cite{voleti2022score}, have also been applied to consider data structures. Correlated noise may be a potential alternative when correlation exists in a data sample, e.g., when frames of a video are considered \cite{ge2023preserve} as correlations, while other video diffusion models usually maintain the default noise type. Additionally, noise from other distributions is also explored. Gamma distribution \cite{nachmani2021denoising} is another feasible alternative with one more degree of freedom and fits to distributions better. 

Soft corruptions can also be treated as a generalized noise for perturbation. Gaussian blur like the heat equation \cite{rissanen2023generative} is introduced for disentanglement of overall color and shape and smooth interpolation if it is used for image data. Soft corruption can also be manually defined operators like masking \cite{daras2022soft} to perturb data. Such operators also destroy data structures as the aforementioned noise does. This greatly extends the expressive power as a wide variety of operators become available \cite{bansal2022cold}. Table \ref{tab:noise_type} visualizes examples of noise types.

Selecting appropriate types depends on the characteristics of data. Isotropic Gaussian noise is broadly applicable but may overlook the prior knowledge of data structure. Gaussian mixtures are better for data with distinct modes, whereas non-isotropic Gaussian noise considers self-correlation. Soft corruption is effective for known perturbation patterns, as it relies on predefined operators \cite{daras2022soft} to fulfill its flexibility.

\subsection{The Terminal Distribution}\label{sec:terminal_distribution}

Diffusion models are assumed to have zero signal-to-noise ratio (SNR) value for terminal distributions to correctly align diffusion training and inference \cite{lin2024common}. Ideally, $x_T$ is heavily perturbed without any original structures, i.e., zero SNR. However, empirically the terminal distribution may not strictly be zero SNR. This misalignment with assumption leads to suboptimal generation quality.

\begin{figure}[tbph]
  \centering
  \includegraphics[width=0.8\linewidth]{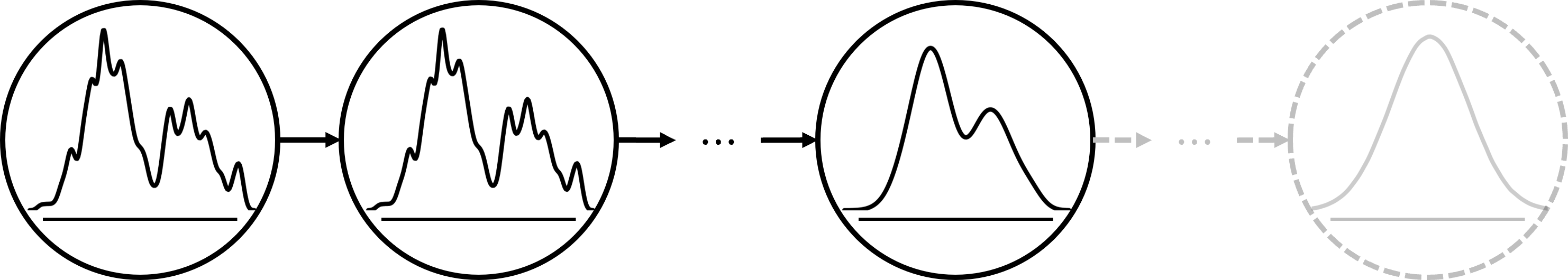}
  \caption{\label{fig:informative_terminal}%
          The transition chain no longer seeks an isotropic Gaussian distribution as the terminal distribution. The grey, dashed parts represent that the transition no longer approaches the isotropic Gaussian distribution.
          }
\end{figure}

The forward process seeks suitable terminal distributions, either to maintain as many structures of $x_0$ as possible or to ensure zero SNR at the terminal distribution. Fig. \ref{fig:informative_terminal} demonstrates the first direction. These approaches usually consider the statistics of the training dataset \cite{lee2021priorgrad} such as the mean and variance to serve as proxies for data structures. Learning the terminal distribution $p(x_T)$ with additional networks is also feasible \cite{zheng2023truncated}. This direction expects that retaining more meaningful structural information from the data distribution mitigates the difficulty of generation. The other direction adjusts the forward process to ensure that the terminal distribution conforms to our assumption. Offset noise is straightforward by altering its mean value but is not devoid of inherent challenges. \cite{lin2024common} rescales the noise schedule but requires subsequent fine-tuning across the entire network. \cite{hu2024one} relies on training an auxiliary text-conditional network to map pure Gaussian noise to the data-adulterated noise.

Choosing a suitable terminal distribution remains an open and important problem. The direction of maintaining more original structures may additionally accelerate the sampling process because fewer timesteps are involved, but may require an accurate representation of the terminal distribution \cite{lee2021priorgrad}. On the contrary, ensuring zero SNR sticks to the assumption of diffusion models and the terminal distribution is accessible, which is more desirable when accurate representations are hard to obtain.

\subsection{Representation Space}\label{sec:data_space}

\begin{figure}[tbph]
  \centering
  \includegraphics[width=0.8\linewidth]{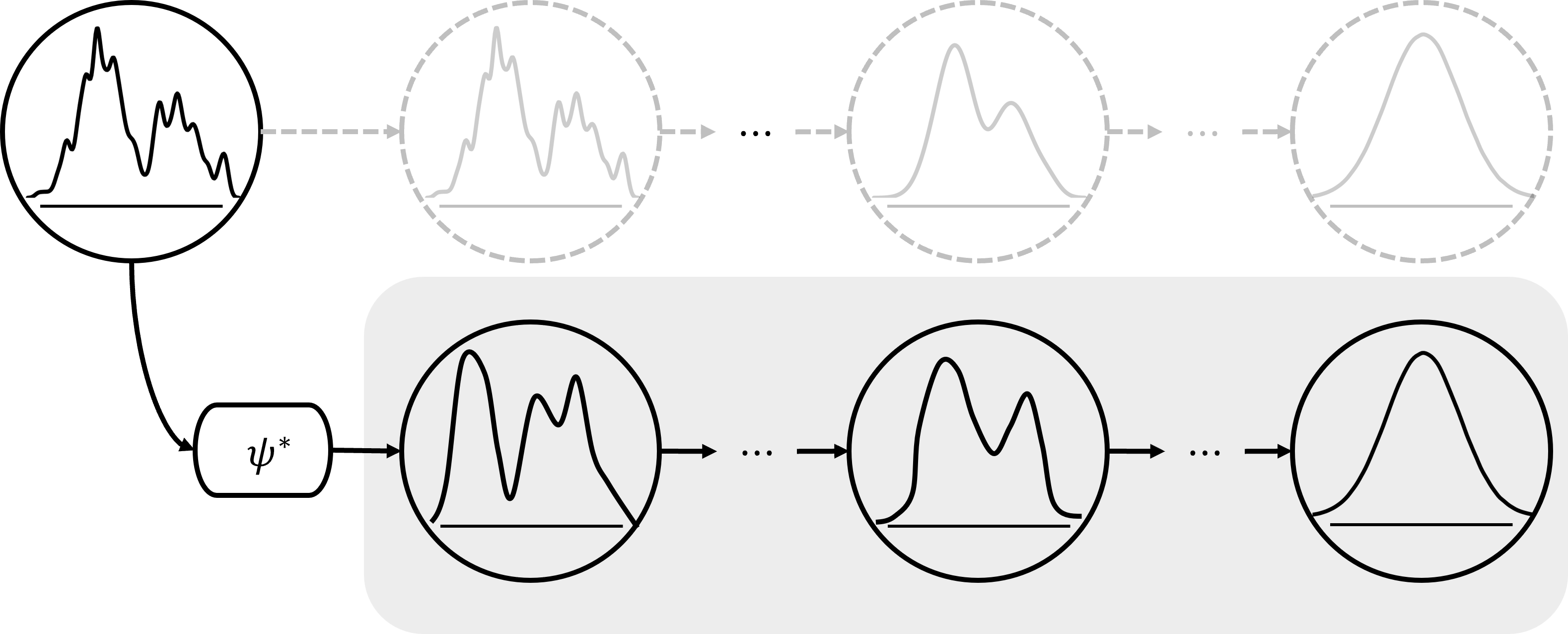}
  \caption{\label{fig:latent_space}%
           The transition chain in a latent space. $\psi^*$ is a pre-trained encoder. Data are no longer manipulated in the original space (dashed, grey). They are now transformed within the latent space (rounded rectangle).
           }
\end{figure}

Latent representation now becomes a common choice for diffusion, as illustrated in Fig. \ref{fig:latent_space}. The high dimensionality of data often leads to considerable computational cost and redundancy. One of the representative models is Latent Diffusion Model \cite{rombach2022high} where images are compressed into lower dimensional vectors. Empirical evidence \cite{pandey2022diffusevae} shows that some transitions in diffusion models are responsible for learning latent representations, which are usually in low-dimensional space and semantically meaningful.

\begin{figure}[tbph]
  \centering
  \includegraphics[width=0.8\linewidth]{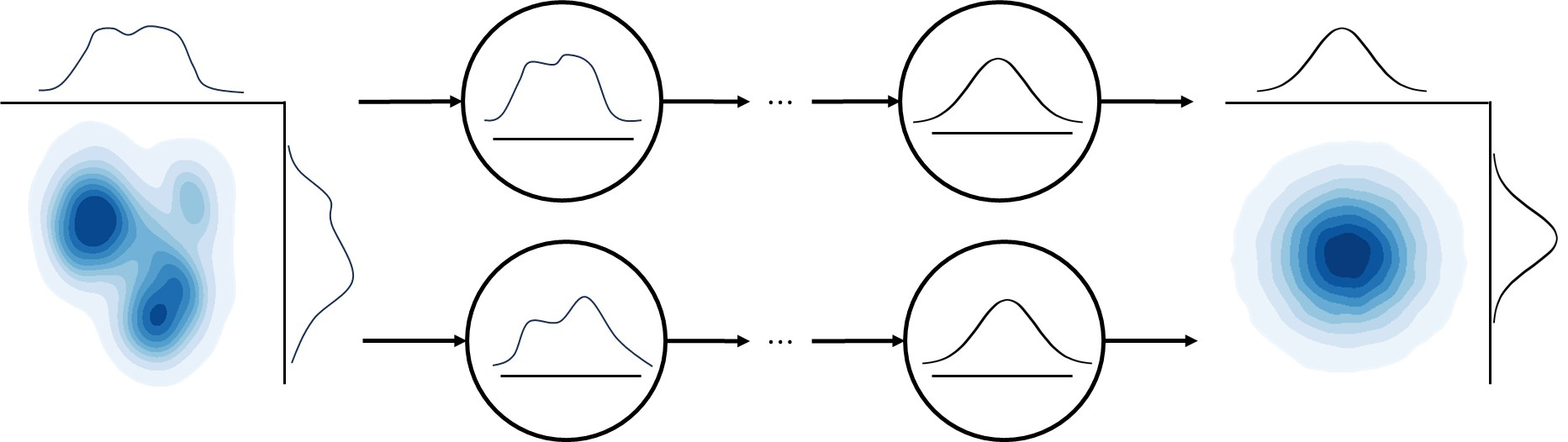}
  \caption{\label{fig:systematic_perspective}%
           The forward process to separately transform the original data in orthogonal subspace.
           }
\end{figure}

Sub-space methods treat different parts of input separately in corresponding sub-spaces. They usually involve more than one chain of transitions in the generic pipeline. In the conditional case, data and their labels are taken as two sub-spaces and then are diffused simultaneously \cite{song2020score}. This design explicitly builds a joint modeling approach for conditional data. Data orthogonal decomposition is widely used where data are decomposed into several complementary parts in their corresponding sub-spaces \cite{choi2023dddmvc}. This design brings flexibility for modeling data with heterogeneous properties. Fig. \ref{fig:systematic_perspective} shows an example of defining sub-spaces by data decomposition.

Augmented space introduces intermediate variables to extend the original space. Introducing intermediate variables is motivated by the overly simplistic diffusions that cannot represent the full dynamics in the forward process \cite{dockhorn2022score}, and thus leads to unnecessarily complex denoising processes and limits generative modeling performance. An auxiliary velocity variable is introduced and the forward process is only defined in the augmented space \cite{chen2024generative}. Introducing stochasticity into the augmented space benefits the smoothness of the evolution of variables.

The three spaces can be mutually beneficial instead of exclusive. Latent representation is more suitable for compressing high-dimensional data but may risk information loss. Sub-space methods offer flexibility for modeling heterogeneous data but add pipeline complexity. Augmented space approaches capture indirect dynamics by introducing intermediate variables.

\section{The Reverse Process}\label{sec:reverse}

The reverse process focuses on training a denoising network to remove noise. The denoising network is configured by its network architecture and its output parameterizations. To train the configured network, optimization designs are also developed.

\subsection{Network Architectures}\label{sec:denoising_architecture}

\subsubsection{Architecture Flexibility}

Theoretically, it is feasible to incorporate as a denoising network a wide variety of architectures that keep the dimensionality unchanged \cite{lee2023multi}. Both U-Net and Transformer have become the mainstream for their high capacity for modeling complex relationships in a wide range of applications and GNN quickly gets more attention. While other architectures may also theoretically be compatible without changing dimensions like GAN \cite{goodfellow2020generative}, they are often adopted for task-specific purposes, e.g., adopting GAN for fast generation \cite{xiao2021tackling}, and may not be generally applicable to other purposes.

\subsubsection{U-Net}

Since \cite{ho2020denoising} first introduced diffusion models with U-Net architecture, this architecture has dominated this area and largely remained intact. From a theoretical standpoint, it is a U-shaped encoder-decoder architecture for general purposes. Its encoder extracts high-level features from data and usually contains downsampling layers to compress data. Its decoder leverages such features for different purposes and usually upsamples back to the original dimensionality of the data. This architecture forms an information bottleneck \cite{tishby2015deep} and encourages the network to learn features effectively. Nonetheless, a few modifications have been introduced in some representative U-Net-based diffusion models such as ADM \cite{dhariwal2021diffusion} and EDM \cite{karras2022elucidating,karras2024analyzing} when compared with the traditional U-Net. \cite{dhariwal2021diffusion} ablates several architecture choices such as adaptive group normalization. Based on \cite{karras2022elucidating}, \cite{karras2024analyzing} proposes magnitude-preserving layers to replace the data-dependent group normalization layers to preserve activation, weight, and update magnitudes. Additionally, the U-shape architecture has been adopted with cross-attention \cite{chen2021crossvit} for higher capacity, and cascading for hierarchy modeling \cite{ho2022cascaded} has been merged into the U-Net architecture. Latent diffusion models \cite{rombach2022high} enhance the traditional U-Net architecture by transformer layers to capture long dependency.

\subsubsection{Transformer}

Transformers are increasingly adapted as an alternative architecture for the denoising network for their superior properties like global dependency, scalability, and multimodality \cite{peebles2023scalable} because of self-attention functions \cite{vaswani2017attention}. In principle, a transformer can directly substitute U-Nets because it can also maintain the data dimensions \cite{lu2023vdt}. However, transformers also exhibit unique advantages. Scalability \cite{peebles2023scalable} of transformers enables better generation quality with less network complexity, which is critical for emergence ability. Besides, as a sequence model, transformers support an arbitrary length of generation. For example, MDM \cite{tevet2022human,chang2025largescale} generates an arbitrary length of human motions. Transformers are also friendly for multi-modality in diffusion models. Stable Diffusion 3 \cite{esser2024scaling} enables the alignment of conditions with MMDIT \cite{reuss2024multimodal} that employs shared full attention weights for visual and textual modalities.

Transformers natively support diverse conditioning via their core mechanisms such as layer normalization and multi-head attention \cite{peebles2023scalable}. Adaptive layer norm modulates features via scale-and-shift. Cross-attention allows each token to selectively focus on condition embeddings for precise control. In-context conditional tokens embed guidance into the representation space. Convolutional U-Nets, in contrast, usually need to graft carefully designed layers or attention blocks at each stage, complicating design and scaling. As a unified framework, transformers streamline implementation and scale predictably, boosting both generative capacity and conditioning fidelity.

Since DiT \cite{peebles2023scalable} demonstrates the superiority of diffusion transformers, different ways to integrate designs from other architectures into transformers have been explored. U-ViT \cite{bao2023all} which utilizes U-Net architecture for diffusion transformers. Mask-DiT \cite{zheng2024fast} and MDT \cite{gao2023masked} use the MAE-like architecture. FiT \cite{lu2024fit,wang2024fitv2} proposes flexible transformer architecture. SANA \cite{xie2024sana} with linear transformer architecture.

\subsubsection{GNN}
Graph Neural Networks (GNNs) have also been an emerging choice for the denoising network especially when graph structures are involved. Their superior performance is attributed to inductive bias from network architecture \cite{zhang2023surveygraph}. Otherwise, the learning may deviate from inherent graph properties \cite{xu2022geodiff}. Equivariant property has been widely considered for denoising architecture. \cite{hoogeboom2022equivariant} keeps the property of invariant permutation in the reverse process. \cite{bao2022equivariant} further extends it with equivariant energy guidance to learn the geometric symmetry. Another considered aspect is the formation of graph structure. \cite{jo2022score} defines the denoising on the adjacency matrix as well as node features. \cite{luo2023fast} performs low-rank Gaussian noise insertion with spectral decomposition. Latent space has also been considered. \cite{yang2023scorebased} encodes the high-dimensional discrete space to low-dimensional topology-injected latent space.

\subsection{Network Parameterizations}\label{sec:network_parameterization}

The output of the denoising network is applied to parameterize the reverse mean $\mu_\theta(x_t,t)$ in the reverse transition. Different parameterization ways all center on the estimation of the original data $x_0$. Specifically, the true value of the reverse mean, denoted as $\mu(x_t,t)$, is formulated as:
\begin{equation}\label{eq:true_reverse_mean}
    \mu(x_t,t):=\frac{\sqrt{\bar{\alpha}_{t-1}}(1-\bar{\alpha}_{t-1})x_t+\sqrt{\bar{\alpha}_{t-1}}(1-\alpha_t){x}_0}{1-\bar{\alpha}_t},
\end{equation}
where ${x}_0$ is the original data but is unavailable during the reverse process. Therefore, $x_0$ needs to be estimated from the observed perturbed data $x_t$ and timestep $t$ by the network. One parameterization way is to directly output the estimation $\hat{x}_0$ by the denoising network and replace $x_0$ with $\hat{x}_0$ in Eq. (\ref{eq:true_reverse_mean}). An indirect parameterization way designs the denoising network to predict the noise $\hat{\epsilon}_t$, which is the residual between the unknown $x_0$ and the observed $x_t$ \cite{ho2020denoising}. Another indirect parameterization way is based on the probabilistic viewpoint and predicts the score $\hat{s}_t$ via the denoising network. $\hat{s}_t$ is the gradient that points towards the unknown $x_0$ from the current position $x_t$ in data space. Combinations among the aforementioned ways are also proposed for special tasks. Table \ref{tab:network_prediction} shows a comparison of different parameterization ways. Different outputs are equivalent to each other \cite{kingma2021variational,karras2022elucidating} with the following relationships:
\begin{align}
    x_\theta(x_t,t)&=\frac{x_t+(1-\bar{\alpha}_t)s_\theta(x_t,t)}{\sqrt{\bar{\alpha}_t}}=\frac{x_t-\sqrt{1-\bar{\alpha}_t}\epsilon_\theta(x_t,t)}{\sqrt{\bar{\alpha}_t}},\\
    s_\theta(x_t,t)&=-\frac{1}{\sqrt{1-\bar{\alpha}_t}}\epsilon_\theta(x_t,t).\\
    \nonumber
\end{align}

While essentially equivalent, different outputs as well as corresponding parameterizations show unique characteristics in particular aspects. Using $\hat{x}_0$ mainly supports better accuracy in the initial stage of the reverse process while $\hat{\epsilon}_t$ is preferable in the late stage. Employing $\hat{s}_t$ avoids computing the normalizing constant, which is a common problem in the context of distribution modeling. Combining the aforementioned ones provides the flexibility to retain their benefits.

\subsubsection{Starting Data}
Predicting the original data $x_0$ provides a straightforward denoising direction. $\hat{x}_0$ indicates a denoising goal towards which $x_t$ should be changed. In particular, given the observation $x_t$ at timestep $t$, the parameterization is defined as:
\begin{equation}\label{eq:predict_data}
    \mu_\theta(x_t,t):=\frac{\sqrt{\bar{\alpha}_{t-1}}(1-\bar{\alpha}_{t-1})x_t+\sqrt{\bar{\alpha}_{t-1}}(1-\alpha_t)x_\theta(x_t,t)}{1-\bar{\alpha}_t},
\end{equation}
where $\alpha_t$ indicates the noise level as defined in Section \ref{sec:discrete_formulation}.

\begin{table}[tbph]
\centering
\begin{tabular}{m{0.2\columnwidth}<{\centering}m{0.4\columnwidth}<{\centering}m{0.25\columnwidth}<{\centering}}
\hline
Output &Parameterization &Visualization \\
\hline
Data $x_\theta$          & $\mu_\theta(x_t,t):=\frac{\sqrt{\bar{\alpha}_{t-1}}(1-\bar{\alpha}_{t-1})x_t+\sqrt{\bar{\alpha}_{t-1}}(1-\alpha_t)x_\theta}{1-\bar{\alpha}_t}$      
&\begin{minipage}{10mm}
      \includegraphics[width=10mm, height=10mm]{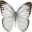}
    \end{minipage}      \\
Score $s_\theta$          & $dx:=\left[f(x,t)-g^2(t)s_\theta\right]dt+g(t)dw$
&\begin{minipage}{10mm}
      \includegraphics[width=10mm, height=10mm]{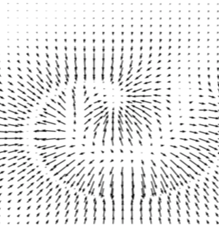}
    \end{minipage}      \\
Noise $\epsilon_\theta$          & $\mu_\theta(x_t,t):=\frac{1}{\sqrt{\alpha_t}}x_t-\frac{1-\alpha_t}{\sqrt{\alpha_t(1-\bar{\alpha}_t)}}\epsilon_\theta$      
&\begin{minipage}{10mm}
      \includegraphics[width=10mm, height=10mm]{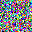}
    \end{minipage}      \\
Hybrid $h_\theta$          & $\mu_\theta(x_t,t):=\mathcal{H}(x_\theta,s_\theta,\epsilon_\theta)$      
& N/A
\\\hline
\end{tabular}
\caption{Visualization of parameterization ways.}
\label{tab:network_prediction}
\end{table}

Parameterizing with $\hat{x}_0$ is advantageous at the beginning of the sampling process, while it leads to inaccuracy when approaching the end of the sampling process. Empirical results show that the estimated mean $\mu_\theta(x_t,t)$, which is parameterized by $\hat{x}_0$, is closer to the ground truth $\mu(x_t,t)$ at the beginning of the sampling process \cite{ramesh2022hierarchical}. This is because $\hat{x}_0$ helps the denoising network with an overall understanding of the global structure \cite{luo2022understanding}. On the contrary, when approaching the end of the sampling process where substantial structures have already been formed and only small noise artifacts need to be removed, finer details are difficult to be recovered \cite{benny2022dynamic}. In other words, the information brought by $\hat{x}_0$ becomes less effectiveness in this case.

\subsubsection{Score}
\begin{figure}[tbph]
  \centering
  \includegraphics[width=0.7\linewidth]{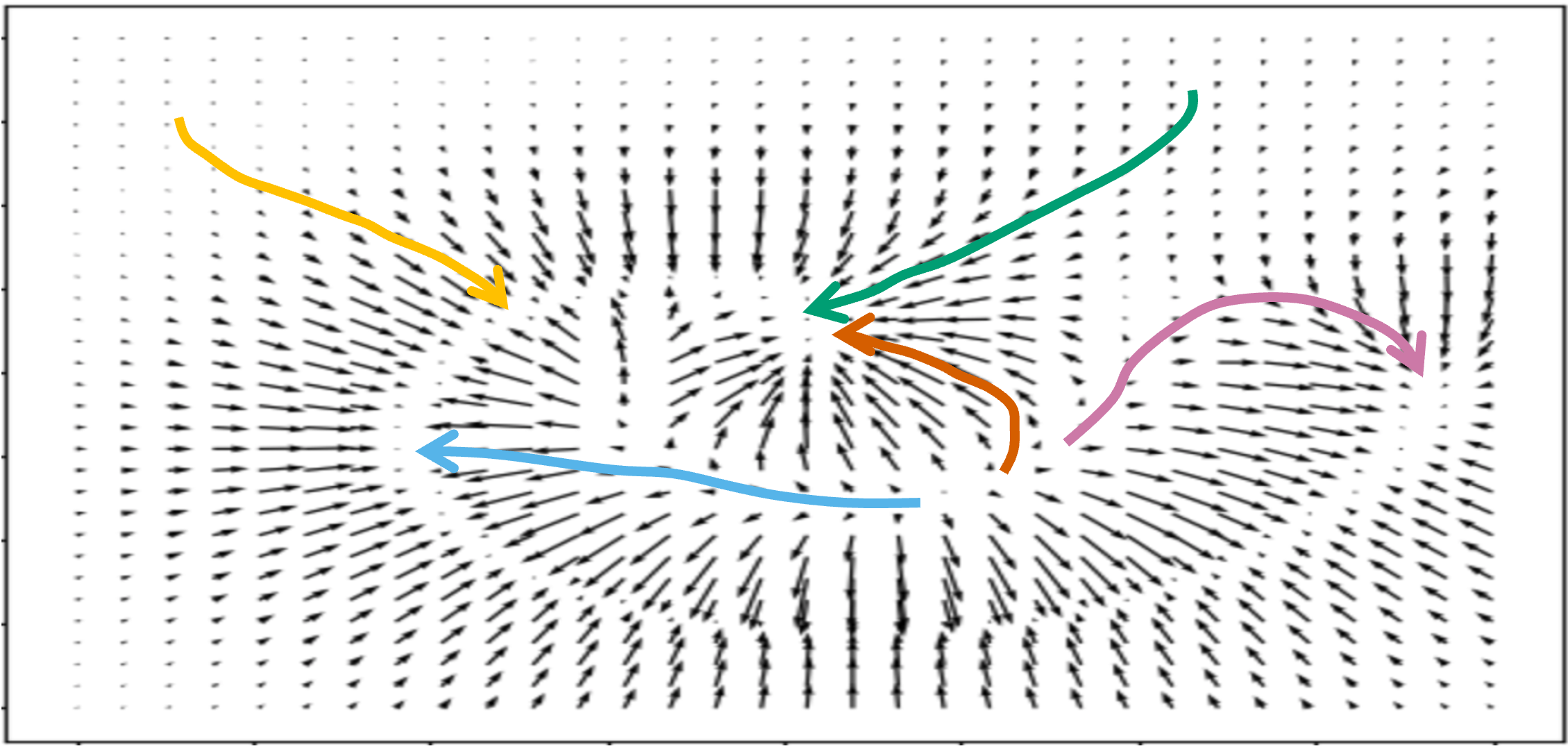}
  \caption{\label{fig:predict_score}%
           Visualization of the trajectory by predicting score. A score is a direction for the next timesteps. Samples are denoised in the direction at each position. Colors represent the trajectories of different samples.
           }
\end{figure}

Score is the gradient of the logarithm of a distribution. The gradient indicates the most possible changes between two timesteps. Therefore, as shown in Fig. \ref{fig:predict_score}, denoising samples by the score forms a trajectory in data space. In particular, given the observed $x_t$ and timestep $t$, the predicted score is defined as $s_\theta(x_t,t):=\nabla_x\log p(x_t)$ and the corresponding parameterization is the reverse SDE:
\begin{equation}\label{eq:predict_score}
    dx:=\left[f(x,t)-g^2(t)s_\theta(x_t,t)\right]dt+g(t)dw,
\end{equation}
where $f(x,t)$ and $g(t)$ are the coefficients as previously introduced in Section \ref{sec:continuous_formulation}.

Score prediction is closely related to flow matching \cite{lipman2023flow} in terms of modeling a vector field. Score-parameterized diffusion models provide unbiased gradients as the vector field under the assumption of Gaussian distribution while flow matching directly learns the vector field. Score-parameterized diffusion models transport data with Gaussian conditional paths while flow matching also supports conditional optimal transport paths. In particular, the predicted distribution is usually defined as:
\begin{equation}
    p_\theta(x)=\frac{\exp^{-f_\theta(x)}}{Z_\theta},
\end{equation}
where $Z_\theta$ is a normalizing constant to estimate. Predicting score avoids this problem:
\begin{align}
        \nabla_x\log p_{t}(x)=-\nabla_xf_\theta(x)-\nabla_x\log Z_\theta=-\nabla_xf_\theta(x),
\end{align}
where $\nabla_x\log Z_\theta=0$ as $Z_\theta$ is a constant with respect to $x$.

\subsubsection{Noise}
\begin{figure}[tbph]
  \centering
  \includegraphics[width=0.75\linewidth]{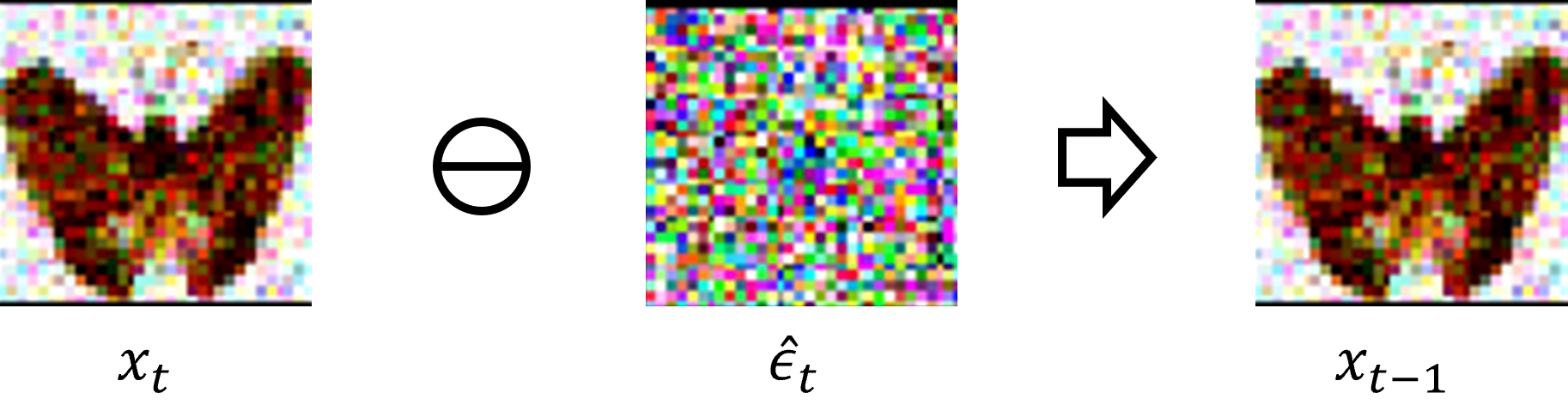}
  \caption{\label{fig:predict_noise}%
           Visualization of the noise-based parameterization. \begin{minipage}{2mm}\includegraphics[width=2mm, height=2mm]{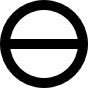}\end{minipage} means $\hat{\epsilon}_t$ has a subtractive relationship with $x_t$, and \begin{minipage}{2mm}\includegraphics[width=2mm, height=2mm]{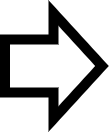}\end{minipage} means this results in $x_{t-1}$.
           }
\end{figure}

Noise estimation predicts the noise added in the forward process. Generally, the predicted noise is scaled according to the noise schedule and then subtracted from the observation \cite{ho2020denoising,xu2022geodiff}, as shown in Fig. \ref{fig:predict_noise}. In particular, given the observation at a current timestep, the prediction of noise is denoted as $\hat{\epsilon}_{t}$ and the parameterization is defined as:
\begin{equation}\label{eq:predict_noise}
    \mu_\theta(x_t,t):=\frac{1}{\sqrt{\alpha_t}}x_t-\frac{1-\alpha_t}{\sqrt{\alpha_t(1-\bar{\alpha}_t)}}\epsilon_\theta(x_t,t),
\end{equation}
where $\alpha_t$ indicates the noise level at timestep $t$ as previously
defined in Section \ref{sec:discrete_formulation}.

The consistent magnitude and residual effect of $\epsilon_\theta(x_t,t)$ are advantageous. The fixed statistics, e.g. $\epsilon_\theta(x_t,t)\sim\mathcal{N}(0,I)$, lead to a consistent magnitude. This encourages the learning of the denoising network \cite{kingma2021variational}. Besides, the residual effect to preserve the input $x_t$ in $x_{t-1}$ is available by predicting zero noise. This becomes increasingly beneficial towards the end of the reverse process where only minor modifications are needed \cite{benny2022dynamic}.

A large deviation between the ground truth noise $\epsilon_t$ and the predicted noise $\epsilon_\theta(x_t,t)$ may occur at the beginning of the sampling process, and is hard to be corrected in the following timesteps. Sampling starts with large noise, with almost no clue for the denoising network to predict noise accurately \cite{ho2020denoising}. This potentially leads to a deviation \cite{benny2022dynamic}. The deviation is scaled up by the noise schedule in Eq. (\ref{eq:predict_noise}). The scheduled level of noise is usually large at the beginning of the sampling process. Even for a small noise estimation error, the deviation will be sharply enlarged. Moreover, the denoising network is limited to predicting noise, which has a residual effect on the noise-based parameterization. The magnitude of potential correction at each timestep is relatively small, and thereby more timesteps are required to correct such a deviation \cite{luo2022understanding}.

\subsubsection{Hybrid}
Combining two or more predictions is also possible for task-specific benefits. Abstractly, the combination is denoted as $h_\theta(x_t,t):=\mathcal{H}(x_\theta(x_t,t),s_\theta(x_t,t),\epsilon_\theta(x_t,t))$ where $\mathcal{H}$ stands for a combination operator. Therefore, the parameterization is:
\begin{equation}\label{eq:predict_combine}
    \mu_\theta(x_t,t)=h_\theta(x_t,t).
\end{equation}
This has a wide variety of feasible implementations because the output to be combined and the combination operators can be very diverse \cite{cao2022survey}. Velocity prediction in DDPM is one example that linearly combines $x_\theta(x_t,t)$ and $\epsilon_\theta(x_t,t)$ \cite{ho2022imagen}, which is designed as:
\begin{equation}
    \mu_\theta(x_t,t):=\alpha_t\epsilon_\theta(x_t,t)-\sigma_tx_\theta(x_t,t),
\end{equation}
where $\alpha_t$ and $\sigma_t$ are the scaling factor and noise schedule respectively. It has better stability \cite{salimans2022progressive}, avoids noise existing in $x_\theta(x_t,t)$ \cite{lin2023common} and achieves higher likelihood \cite{zheng2023improved}. Dynamically alternating between $x_\theta(x_t,t)$ and $\epsilon_\theta(x_t,t)$ accelerates the generation \cite{benny2022dynamic}.

\subsubsection{The Reverse Variance}
Modeling the reverse variance improves the training efficiency of diffusion models. An appropriate variance minimizes the discrepancy between the predicted reverse transition $p_\theta(x_{t-1}|x_t)$ and the forward transition $p(x_t|x_{t-1})$, fitting the forward process better \cite{bao2022analytic}. This facilitates fewer timesteps to be used, and improves overall efficiency.

Many efforts to model the reverse variance are attempted. Some empirically adopt a handcrafted value for each timestep. The noise schedule is a popular option for its simplicity and empirical performance \cite{song2020denoising}. Scaling the schedule by a factor is also researched but does not lead to a large difference \cite{ho2020denoising}. Both choices are considered as upper and lower bounds on reverse process entropy \cite{sohl2015deep}, and the interpolation between them is learned for flexibility \cite{nichol2021improved}. Others find the optimal variance can be solved analytically. Its formulation is explicitly derived from the predicted score \cite{bao2022analytic}, and improves the efficiency of generation \cite{bao2022estimating}.

\subsection{Weighted Optimization}\label{sec:weighted_optimization}

Weighted optimization in the reverse process is inspired by the understanding of the learning procedure of diffusion models. A common choice \cite{ho2020denoising} applies uniform weights and may overlook the characteristics of the reverse process. The semantic information of data expressed in the reverse process gradually changes, which requires appropriately set priorities to learn \cite{stein2024exposing}. An alternative choice is a function of intermediate characteristics, e.g., signal-to-noise values. While it takes data into account, the hyperparameters of the designed function may need to be carefully set.

Learning priorities are balanced by weights in the learning objective to enhance the learning quality. The change of learning priorities has been observed in the reverse process. It pays more attention to global structures at the beginning of the reverse process and then changes to local details when approaching its end \cite{choi2022perception}. A balance is achievable through adjusting weights and beneficial for training.

\begin{figure}[tbph]
  \centering
  \includegraphics[width=0.8\linewidth]{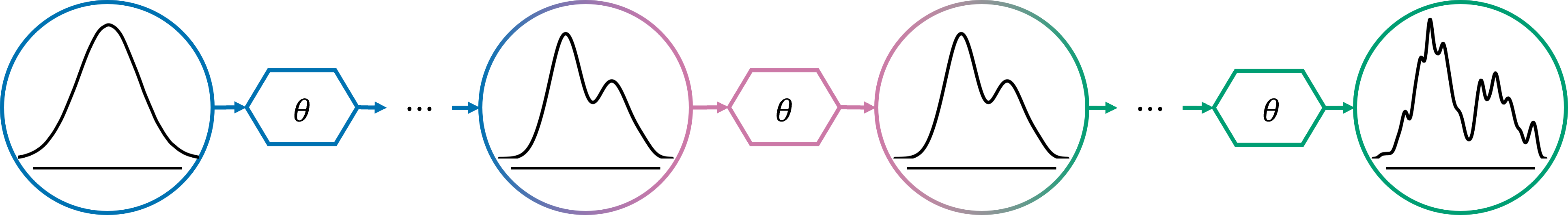}
  \caption{\label{fig:weighting_scheme}%
          The learning priority changes in the reverse process, which is denoted by different colours.
          }
\end{figure}
Directly using the schedule as the weight emphasizes the global structure better by a larger learning weight at the beginning of the reverse process \cite{song2021maximum}. Despite its simplicity, the pre-defined schedule is not flexible and may deviate away from the actual demands. A function of the noise schedule, such as the signal-to-noise (SNR) ratio, is designed to compute the weights. The actual remaining noise is measured rather than the scheduled one \cite{kingma2021variational}. It takes the data into account, and better balances the learning of local details and global structures \cite{choi2022perception}.

\section{Sampling Process}\label{sec:sampling}

Conditional and fast generation are two focused factors of the sampling process in diffusion models. Without modeling conditions, diffusion models usually do not generate data of high quality when data are considered to follow a conditional distribution \cite{chen2022sampling}. Effective mechanisms of guidance are designed to modify transitions in the sampling process to be compatible with conditions. Moreover, the sampling process is several times slower when compared with other generative models \cite{jolicoeur2021gotta}. The long generation time is mainly attributed to the large number of timesteps. Thus, designs for acceleration are explored to reduce timesteps without heavily impairing quality.

\subsection{Guidance Mechanisms}\label{sec:incorporating_guidance}

Vanilla guidance merges conditions via fusion approaches such as the attention layer. However, the weight of conditions is not easy to adjust. Classifier guidance leverages an additional classifier and adjustable weights, but comes with issues of computational cost and stability. Classifier-free guidance additionally trains unconditional diffusion models to achieve better stability. Learned modifications via adapters provides guidance but need to fine-tune extra adapters.

\subsubsection{Vanilla Guidance}

\begin{figure}[tbph]
  \centering
  \includegraphics[width=0.8\linewidth]{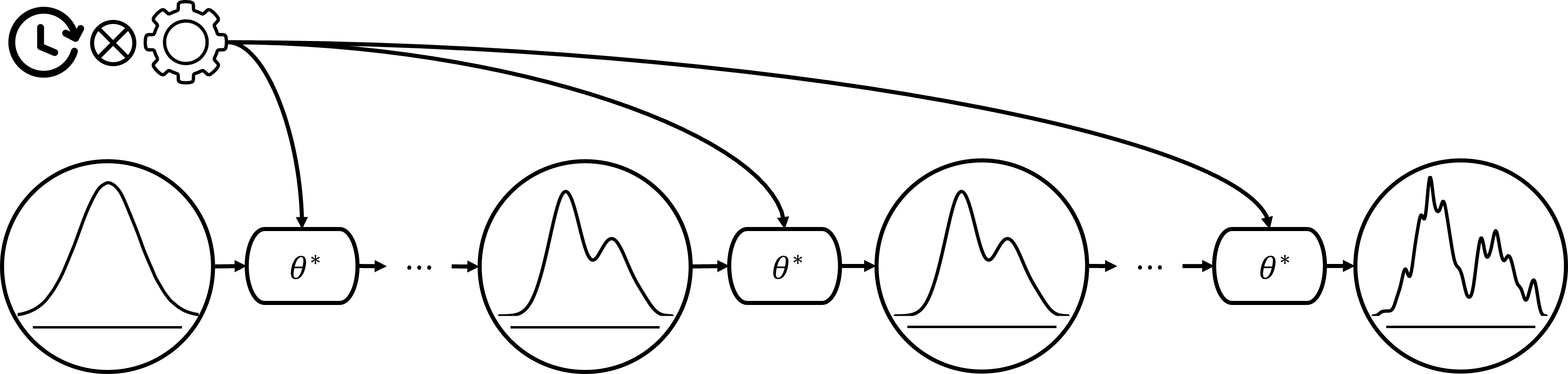}
  \caption{\label{fig:vanilla_guidance}%
          Vanilla guidance adds the conditions $c$ to each timestep $t$ as a new condition.
          }
\end{figure}

Vanilla guidance usually merges the given conditions $c$ with timesteps $t$ as the guidance. The intuition of merging is that a timestep $t$ itself is inherently taken as a condition by a denoising network and thus more conditional information can also be conveyed. The approaches of merging can be operations such as addition \cite{ho2020denoising,chang2022unifying,edmund2022style}, and attention layer \cite{chefer2023attend,hong2023improving}. Fig. \ref{fig:vanilla_guidance} shows the condition is added to a timestep in this mechanism.

While vanilla guidance benefits from its simplicity, its effectiveness is undermined by the lack of adjustable conditional strength \cite{ho2022classifier}. Empirical evidence \cite{luo2022understanding} shows that a conditional diffusion model trained with vanilla guidance may not conform to the conditions or underperform in conditional generation.

\subsubsection{Classifier Guidance}

\begin{figure}[tbph]
  \centering
  \includegraphics[width=0.80\linewidth]{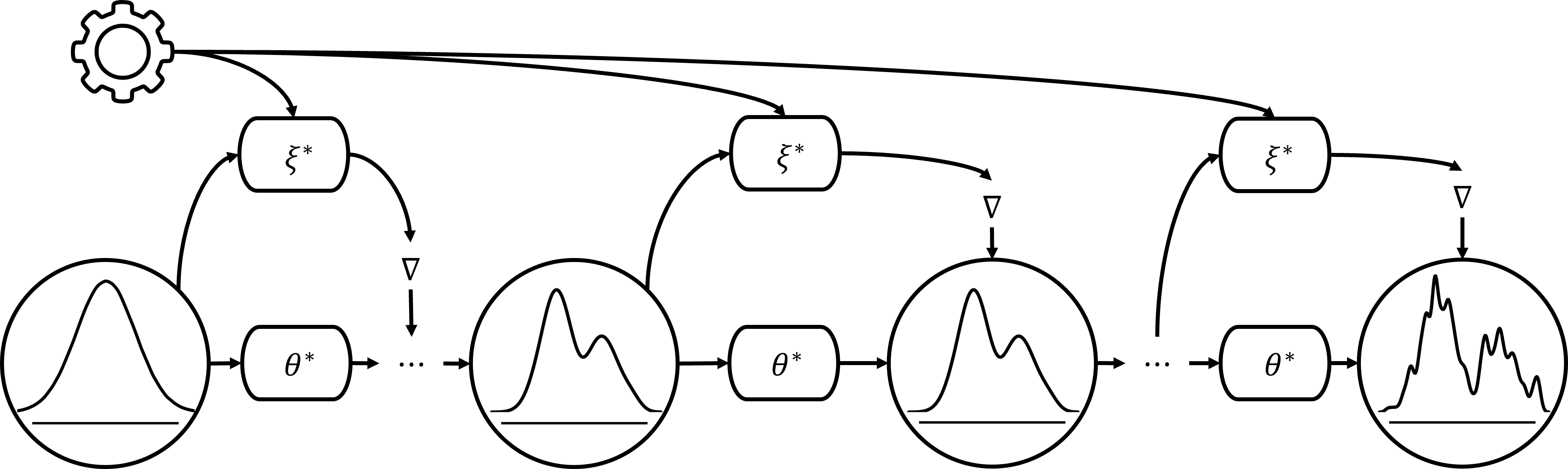}
  \caption{\label{fig:classifier_guidance}%
          Classifier guidance leverages an extra classifier network $\xi^*$ to compute a gradient $\nabla$ as the modification on the denoising network $\theta^*$. The timestep condition $t$ is omitted here for visualization.
          }
\end{figure}

For effective and adjustable strength of conditions, classifier guidance \cite{dhariwal2021diffusion} adopts an extra classifier. The gradient of the classifier is scaled by the weight and then is used to modify the unconditional denoising direction, as shown in Fig. \ref{fig:classifier_guidance}. In other words, the weight controls how much to rely on the classifier. To obtain the gradient as accurately as possible, the classifier is usually pre-trained on data with conditions. In particular, classifier guidance is formulated as:
\begin{equation}\label{eq:classifier_guidance}
    \nabla_x\log p(x|c)=\nabla_x\log p(x)+w\nabla_x\log p(c|x),
\end{equation}
where $\nabla_x\log p(x|c)$ and $\nabla_x\log p(x)$ are conditional and unconditional scores, respectively, $\nabla_x\log p(c|x)$ is the gradient of a classifier, and $w$ is the weight. When $w=0$, this mechanism becomes unconditional. As the weight increases, the denoising network is more and more constrained to produce samples that satisfy conditions.

Additionally learning a classifier may lead to extra cost and training instability. The extra expense is further scaled up because the classifier is trained on data with every scheduled noise level \cite{ho2022classifier}. Moreover, training the classifier on data with noise tends to be unstable. The data structure is almost destroyed because more and larger noise is added according to the noise schedule. Therefore, the quality of the classifier gradient may not be consistent \cite{wallace2023end}. Sometimes its direction is arbitrary or even opposite \cite{srinivas2020rethinking} and leads to less effective or wrong guidance. Some imitate the form of classifier guidance but solve the gradient analytically \cite{yang2024guidance} for the inverse problem to bypass the disadvantages of classifiers.

\subsubsection{Classifier-Free Guidance}

\begin{figure}[tbph]
  \centering
  \includegraphics[width=0.80\linewidth]{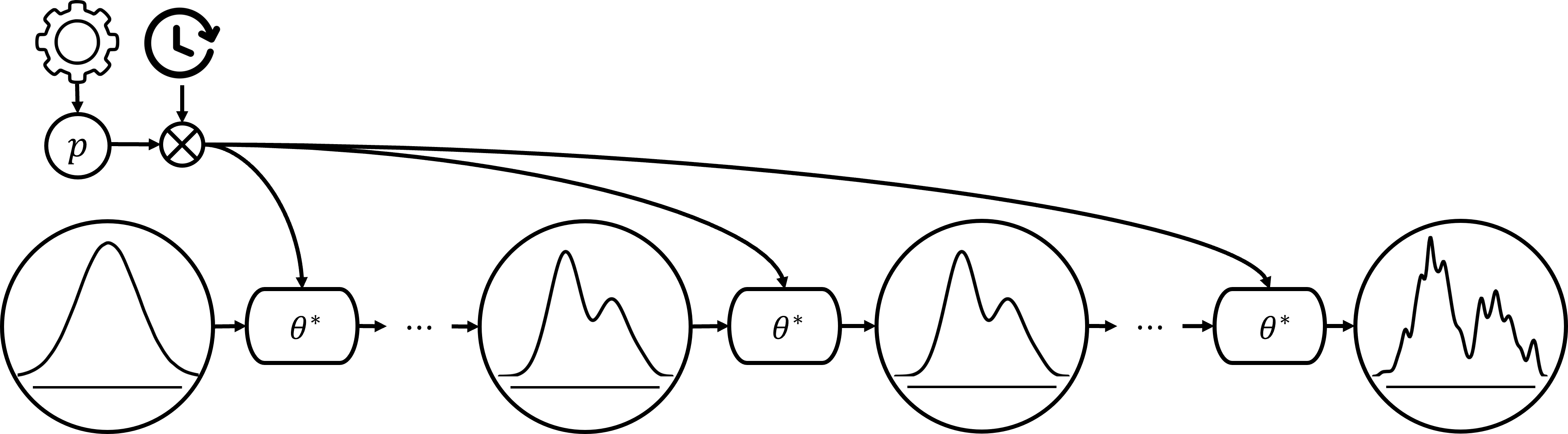}
  \caption{\label{fig:classifier_free}%
          Classifier-free guidance is based on a mixture of vanilla guidance and unconditional model $\theta^*$. A probability $p$ controls whether to drop out the conditions during training.
          }
\end{figure}

To avoid the extra classifier, classifier-free guidance \cite{ho2022classifier} replaces the classifier with a mixture of unconditional and condition models. It further enhances the sampling process to follow the direction of guidance by discouraging it away from unconditional direction \cite{zhao2023null}. As shown in Fig. \ref{fig:classifier_free}, instead of just training a conditional model, an unconditional one is also trained simultaneously by dropping out conditions $c$ with a probability $p$. In particular, classifier-free guidance is formulated as:
\begin{equation}
    \nabla_x\log p(x|c)=w\nabla_x\log p(x|c)+(1-w)\nabla_x\log p(x),
\end{equation}
where $w$ is the weight of conditions. 

The weight is slightly different from its counterpart in classifier guidance. When $w=0$, the classifier-free guidance becomes unconditional models without vanilla guidance. The vanilla guidance is a special case when $w=1$. In this case, the unconditional model is suppressed and conditions are incorporated through vanilla guidance \cite{pang2023calibrating}. If $w>1$, the classifier-free guidance restrains the unconditional model and prioritizes conditions further by larger weights. The score from classifier-free guidance deviates quickly away from the unconditional score, and thus, samples that better satisfy the conditions will be generated \cite{pearce2023imitating}.

\subsubsection{Learned Modifications}
Learning modifications provides greater flexibility for controllable generation by preserving a generative prior. ControlNet \cite{zhang2023adding} is a pioneering method to fine-tune an extra copied model. This idea for conditional generation has been popular, evidenced by various adapters. SUPIR \cite{yu2024scaling} trains a trimmed ControlNet for image restoration. Uni-ControlNet \cite{zhao2023unicontrolnet} further proposes all-in-one control. StableSR \cite{wang2024exploiting} learns a time-aware encoder to assist super-resolution generation that is conditioned on low-resolution images. T2I-Adapter \cite{mou2024t2i} equips the pre-trained model with several low-complexity adapters. LoRAdapter \cite{stracke2024loradapter} takes inspiration from LowRank-Adaptations (LoRA) \cite{hu2022lora} to further reduce the parameters to be learned.

\begin{figure}[tbph]
  \centering
  \includegraphics[width=0.8\linewidth]{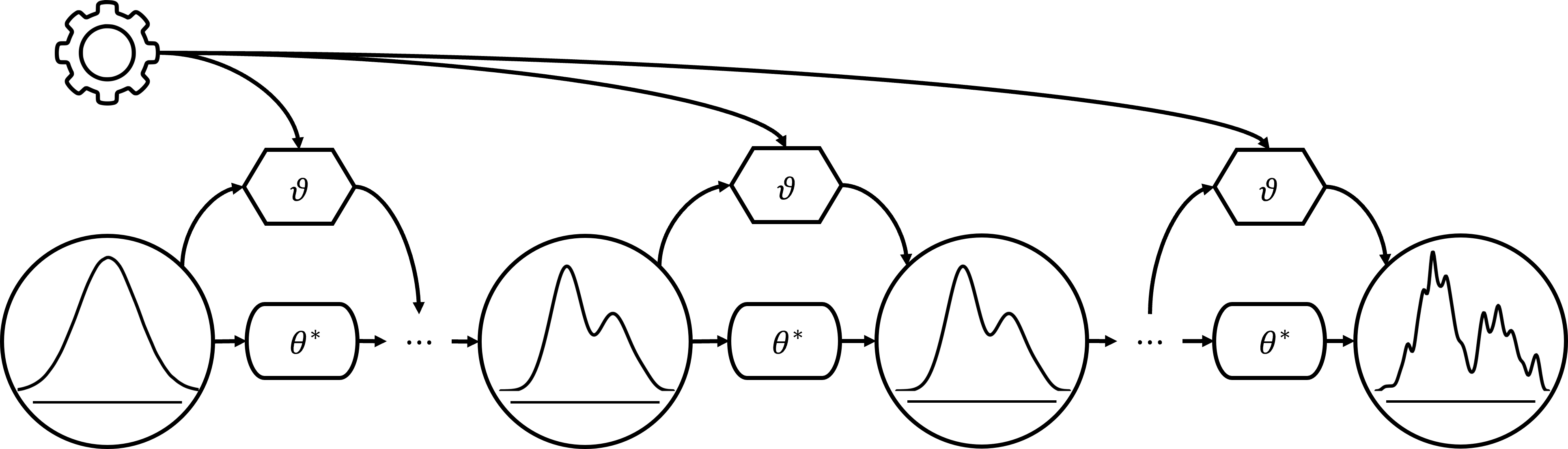}
  \caption{\label{fig:learning_guidance_controlnet}%
          Applying an extra network $\vartheta$ to directly learn the required modification for guidance. Timestep condition is omitted here.
          }
\end{figure}

\subsection{Acceleration Designs}\label{sec:applying_acceleration}

Reducing the number of timesteps for generation is the main goal of acceleration. Generally, the denoising network needs to wait for the results from the timestep $t+1$ to accomplish the transition at the current timestep $t$. The inference speed is significantly slowed down especially when a large number of timesteps are required in the sampling process \cite{xiao2021tackling}. Truncation directly cuts the sampling process at a certain timestep but may suffer from distribution deviation. Knowledge distillation is adopted to learn an auxiliary module that enables fewer timesteps in its sampling process but may require careful fine-tuning. Timestep selection skips some timesteps and selects a subset of timesteps for fast generation but may suffer from inherent approximation errors.

\subsubsection{Truncation}

\begin{figure}[tbph]
  \centering
  \includegraphics[width=0.8\linewidth]{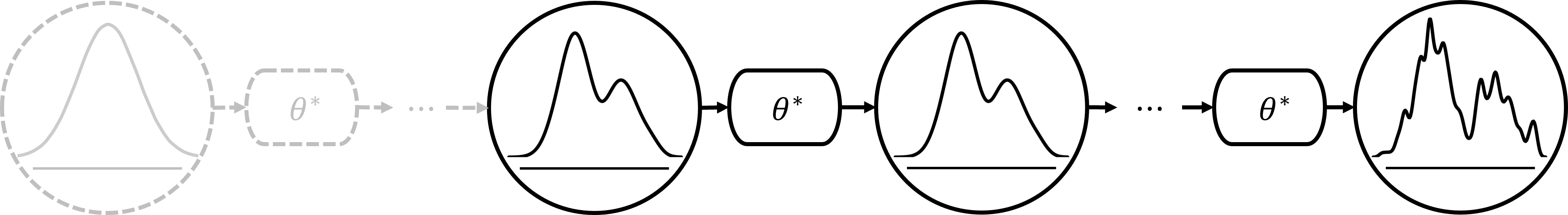}
  \caption{\label{fig:truncated_sampling}%
          The sampling process is truncated and starts from a selected timestep. The grey, dashed parts represent discarding for generations.
          }
\end{figure}

Truncation involves a partial sampling process with an extra network. It usually selects an intermediate timestep $t^\prime$, and obtains a sample from the corresponding distribution $p(x_{t^\prime})$ for the generation, as shown in Fig. \ref{fig:truncated_sampling}. In other words, the process truncates the whole chain at $t^\prime$, and thereby fewer timesteps remain in the partial chain. An extra network needs to be additionally trained to model $p(t^\prime)$ that may not be tractable \cite{ho2020denoising}. Overall, truncation is theoretically effective in acceleration \cite{chung2022come}, which is proved by the stochastic contraction theory.

Truncation effects can be two-sided. On the one hand, truncation comes with acceleration for not only inference but also training. Truncation also strikes a balance between acceleration and quality as the selection of such a point depends on the data complexity \cite{zheng2023truncated} and the degree of corruption \cite{chung2022come}. Besides, truncation takes advantage of the properties of the involved extra network, which is often another generative model \cite{lyu2022accelerating}. On the other hand, truncation may lead to an increased training expense \cite{lyu2022accelerating} because the extra network needs to learn $p(t^\prime)$ as accurately as possible.

\subsubsection{Knowledge Distillation}

\begin{figure}[tbph]
  \centering
  \includegraphics[width=0.7\linewidth]{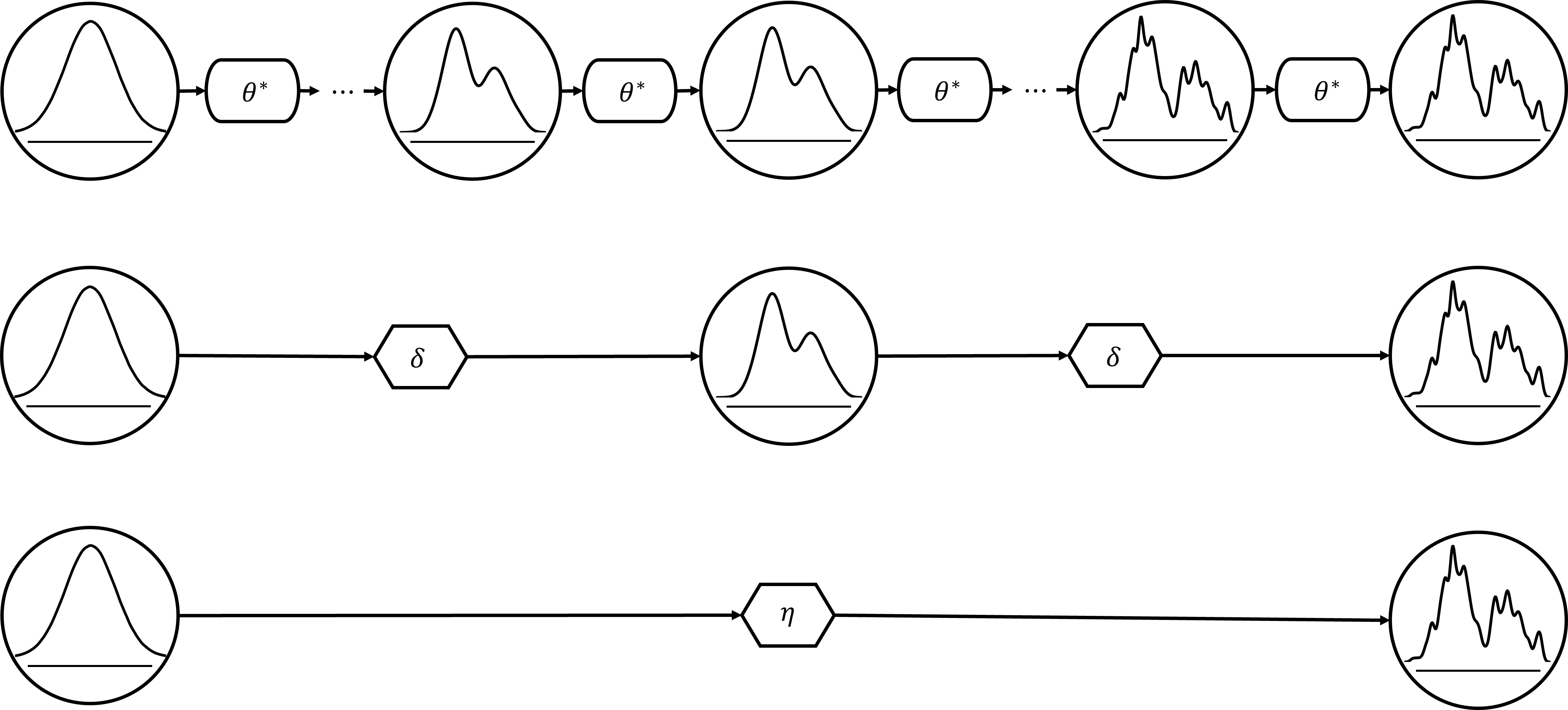}
  \caption{\label{fig:knowledge_distillation}%
          Knowledge distillation learns student denoising networks $\delta$ and $\eta$ with fewer timesteps, based on the teacher denoising network $\theta^*$.
          }
\end{figure}

The technique can also be applied to learn a new sampling process with fewer timesteps. Knowledge distillation is a network compression technique. In terms of diffusion models, it involves the original sampling process as the teacher model and a new one with fewer timesteps as the student model \cite{huang2023knowledge}. Fig. \ref{fig:knowledge_distillation} shows an example method of progressive distillation in the sampling process.

Knowledge distillation is usually applied to merge several timesteps into fewer timesteps for the new sampling process such as progressive distillation, guidance distillation, and consistency models. \cite{luhman2021knowledge} directly distills all timesteps into a single one with expensive computation since a large dataset of samples from the teacher model is needed \cite{meng2022distillation}. One mitigation is progressively reducing timesteps \cite{salimans2022progressive}. Guidance distillation \cite{hsiao2024plug} is a plug-and-play acceleration approach with classifier guidance but the memory usage is high. Another family is consistency models where the sampling trajectory of diffusion models are straightened for consistency \cite{song2023consistency} to accelerate the sampling process. Straightening is achieved via two approaches. Consistency training approach learns a consistency model from scratch, as if it distilled all timesteps into a single timestep to achieve consistency. In contrast, consistency distillation accelerates existing models to become consistency models for single or fewer timesteps.

\subsubsection{Timestep Selection}

\begin{figure}[tbph]
  \centering
  \includegraphics[width=0.8\linewidth]{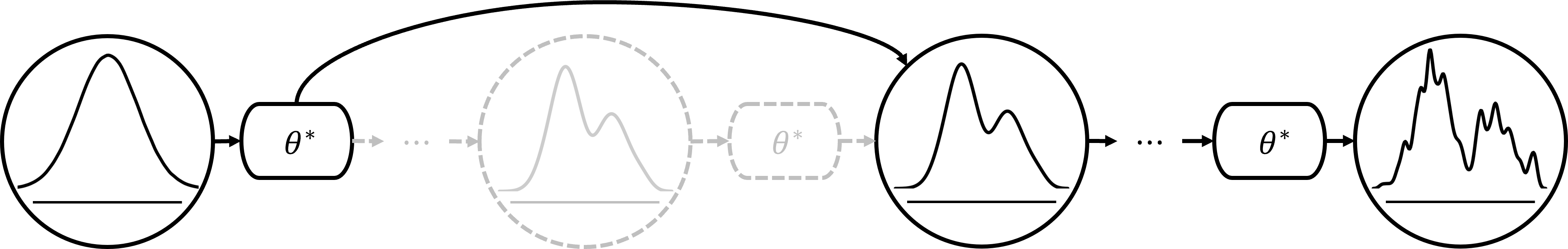}
  \caption{\label{fig:sampling_solver}%
          Selection strategies for the sampling process skip the selected time steps for generation.
          }
\end{figure}

Timestep selection seeks to develop strategies to select only a subset of timesteps without undermining quality. Some timesteps in a sampling process influence quality less and thus they can be skipped safely \cite{fang2023structural}. Fig. \ref{fig:sampling_solver} shows a shorter sampling process with selected timesteps. Some may not directly reduce the number of timesteps but select a subset of timesteps for parallel computation \cite{shih2023parallel}.

Some acceleration methods essentially solve stochastic or ordinary differential equations. DDIM \cite{song2020denoising} has been widely applied for its simplicity in using non-Markovian reverse processes, and PNDM \cite{liu2022pseudo} further builds close connections with pseudo-numerical methods for acceleration. Another commonly used method is DPM-Solver \cite{lu2022dpm} and its improved version DPM-Solver++ \cite{lu2023dpmsolver}. They are high-order solvers for ODEs with a convergence order guarantee. Empirically, 20 timesteps are used to generate high-quality results. Euler samplers are also widely used \cite{karras2022elucidating} to generate with 20 to 30 timesteps. The selection of timesteps can also be learned by extra networks \cite{duan2023optimal} or rely on dynamic programming algorithms \cite{watson2021learning}.

\section{Future Trends}\label{sec:future}

We provide a summary in Table \ref{tab:summary_table} to connect the components in this survey with popular open-source diffusion models.

\begin{table}[htbp]
\scriptsize
\centering
\begin{tabular}{cccc}
\toprule
Models              & Forward & Reverse & Sampling                 \\
\toprule
\multirow{3}{*}{DiT \cite{peebles2023scalable}}        & Schedule: Linear      & Architecture: Transformer     & \multirow{2}{*}{Guidance:} \multirow{2}{*}{Vanilla/CFG}  \\
 & Type: Gaussian & Parametrization: Noise     &                    \\
 & Space: Latent & Weight: Pre-defined & Acceleration: Solvers \\
 \midrule
\multirow{3}{*}{SD1 \cite{rombach2022high}}         & Schedule: Scaled Linear      & Architecture: U-Net           & \multirow{2}{*}{Guidance:} \multirow{2}{*}{Vanilla/CFG}  \\
 & Type: Gaussian     & Parametrization: Noise      &                    \\
 & Space: Latent & Weight: Pre-Defined & Acceleration: Solvers \\
 \midrule
\multirow{3}{*}{SD2 \cite{stablediffusion2}}        & Schedule: Scaled Linear      & Architecture: U-Net           & \multirow{2}{*}{Guidance:} \multirow{2}{*}{Vanilla/CFG}  \\
 & Type: Gaussian     & Parametrization: Hybrid (v)      &                    \\
 & Space: Latent & Weight: Pre-Defined & Acceleration: Solvers \\
 \midrule
\multirow{3}{*}{SD3 \cite{esser2024scaling}}        & Schedule: Rectified Flow      & Architecture: Transformer     & \multirow{2}{*}{Guidance:} \multirow{2}{*}{Vanilla/CFG}  \\
 & Type: Gaussian     & Parametrization: Score (flow)      &                    \\
 & Space: Latent & Weight: Pre-Defined & Acceleration: Solvers \\
 \midrule
\multirow{3}{*}{SVD \cite{blattmann2023stable}}        & Schedule: Linear      & Architecture: U-Net           & \multirow{2}{*}{Guidance:} \multirow{2}{*}{Vanilla/CFG} \\
 & Type: Gaussian     & Parametrization: Hybrid (v)      &                    \\
 & Space: Latent & Weight: Pre-Defined & Acceleration: Solvers \\
 \midrule
\multirow{3}{*}{OpenSora1.2 \cite{opensora}}   & Schedule: Rectified Flow      & Architecture: Transformer     & \multirow{2}{*}{Guidance:} \multirow{2}{*}{CFG} \\
 & Type: Gaussian     & Parametrization: Score (flow)     &                    \\
 & Space: Latent & Weight: Pre-Defined & Acceleration: Solvers \\
 \midrule
\multirow{3}{*}{FLUX \cite{flux}}       & Schedule: Rectified Flow      & Architecture: Transformer     & \multirow{2}{*}{Guidance:} \multirow{2}{*}{Vanilla/LoRA/CFG} \\
 & Type: Gaussian     & Parametrization: Score (flow)      &                    \\
 & Spcae: Latent & Weight: Pre-Defined & Acceleration: Distillation      \\
 \midrule
\end{tabular}
\caption{Popular diffusion models with their designs. Guidance can vary across tasks because of the versatility and adaptability of diffusion models.}
\label{tab:summary_table}
\end{table}

\subsection{Generalization Capability}
Understanding the reasons why diffusion models generalize well remains an open problem. Recent advances have attributed the generalization to a few designs such as the Gaussian structure \cite{li2024understanding} and geometry-adaptive harmonic representations \cite{kadkhodaie2024generalization}, while the comprehensive underlying mechanism remains unclear. Further investigation can explore other designs such as parameterizations. This may need to further consider the interplay effects of other designs with extensive analysis and experiments since designs are often deeply correlated.

\subsection{Denoising-Oriented Architecture}

The network architectures of diffusion models present significant research opportunities. Section \ref{sec:denoising_architecture} showcases success through integration with other areas. Nonetheless, a lack of specialized denoising architectures may result in suboptimal denoising performance. Future trends may borrow network architectures from related fields like image restoration \cite{jiang2024survey}, which also aim to recover unperturbed data from noisy inputs. Adaptation to the wide variety of degradations may require additional modifications to the forward and reverse processes. Additionally, timestep-adaptive architectures may also be considered to better leverage the inherent denoising priority \cite{lee2023multi} in diffusion processes. However, adopting timestep-adaptive architectures increases the computational and architectural complexity, requiring additional efforts to balance the tradeoffs.

\subsection{Responsible Applications}

Diffusion models have become popular in sciences, such as physics \cite{shmakov2023end} and medicine \cite{gou2024cascaded}, etc. Nonetheless, concerns remain on whether they are reliable to capture the underlying causal relations of a domain as diffusion models tend to rely on statistical associations on the training dataset \cite{lorch2024causal}. One possible direction may integrate causality-aware guidance mechanisms to condition diffusion models on directed causal graphs. While promising, handling unobserved confounding relationships usually requires extensive assumption tests to achieve higher reliability \cite{chao2023interventional}.

Diffusion models such as Stable Diffusion \cite{esser2024scaling} have been significantly employed in the creative industry. Nonetheless, concerns about originality in creative works have been raised \cite{gu2023memorization} and generated content may infringe copyrights. One potential direction adopts concept forgetting \cite{zhang2024forget} or adjusts the sample process with likelihood-aware guidance \cite{chen2024towards} to reduce the similarity. The quantification of originality remains an open problem for creative industries \cite{erickson2018intellectual}, and establishing industry standards to generate content ethically requires broad industry cooperation.

\subsection{Societal Impacts}

Diffusion models showcase biases, and critically, the definition of biases evolves over time and varies across cultures. Current solutions do not consider this evolution and variation. Combining fairness-aware algorithms \cite{choi2024fair} with incremental learning \cite{gao2023ddgr} may facilitate “incremental fairness”. The main difficulty lies in the interdependent and multifaceted nature of biases where additional data may give rise to new biases in existing data and addressing one type of bias without considering others can result in incomplete or skewed mitigation efforts.

Diffusion models democratize personalization and boost productivity by enabling non-professional users to automatically generate highly customizable content. This accessibility may reshape the job market and further impact educational systems. Consequently, governments and individuals may require additional expenses for employee retraining and educational reforms \cite{zhang2024survey}. Addressing this requires interdisciplinary collaboration among computer science, politics, and economics to promote democratization and productivity while reducing potential social risks \cite{wei2024understanding}.

\section{Conclusion}\label{sec:conclusion}
Diffusion models involve three main components: a forward process and a reverse process for optimization, and a sampling process for generation. The forward process focuses on perturbing data with different noise schedules, noise types, terminal distributions and representations. The reverse process focuses on training a denoising network to remove noise with different architectures, parameterizations, and weights. The sampling process works for generation and mainly focuses on guidance and acceleration. These designs have all contributed to the current powerful diffusion models. Several future trends have been introduced to boost this field.

\setstretch{0.9}
\bibliographystyle{elsarticle-num}
\bibliography{Manuscript}

\end{document}